RESEARCH ARTICLE

# CONCERT: a Modular Reconfigurable Robot for Construction

Luca Rossini*[1,4] | Edoardo Romiti[1,4] | Arturo Laurenzi[1] | Francesco Ruscelli[1] | Marco Ruzzon[1] | Luca Covizzi[1] | Lorenzo Baccelliere[1] | Stefano Carrozzo[1] | Michael Terzer[3] | Marco Magri[3] | Carlo Morganti[3] | Maolin Lei[1] | Liana Bertoni[1] | Diego Vedelago[2] | Corrado Burchielli[1] | Stefano Cordasco[2] | Luca Muratore[1] | Andrea Giusti[3] | Nikos Tsagarakis[1]

[1]Humanoids and Human Centered Mechatronics (HHCM), Istituto Italiano di Tecnologia, Via Morego 30, 16163 Genova, Italy

[2]Advanced Robotics Facility (ADVRF), Istituto Italiano di Tecnologia, Via Morego 30, 16163 Genova, Italy

[3]Fraunhofer Italia Research, Via Alessandro Volta 13a, 39100 Bolzano, Italy

[4]This authors contributed equally to this work.

**Correspondence**

*Luca Rossini, Istituto Italiano di Tecnologia, Via Morego 30, 16163 Genova, Italy.
Email: luca.rossini@iit.it

**Funding Information**

This research was supported by the the Horizon 2020 Research and Innovation Program of the European Union under Project CONCERT (Grant agreement no 101016007)

**Abstract**

This paper presents CONCERT, a fully reconfigurable modular collaborative robot (cobot) for multiple on-site operations in a construction site. CONCERT has been designed to support human activities in construction sites by leveraging two main characteristics: high-power density motors and modularity. In this way, the robot is able to perform a wide range of highly demanding tasks by acting as a co-worker of the human operator or by autonomously executing them following user instructions. Most of its versatility comes from the possibility of rapidly changing its kinematic structure by adding or removing passive or active modules. In this way, the robot can be set up in a vast set of morphologies, consequently changing its workspace and capabilities depending on the task to be executed. In the same way, distal end-effectors can be replaced for the execution of different operations. This paper also includes a full description of the software pipeline employed to automatically discover and deploy the robot morphology. Specifically, depending on the modules installed, the robot updates the kinematic, dynamic, and geometric parameters, taking into account the information embedded in each module. In this way, we demonstrate how the robot can be fully reassembled and made operational in less than ten minutes. We validated the CONCERT robot across different use cases, including drilling, sanding, plastering, and collaborative transportation with obstacle avoidance, all performed in a real construction site scenario. We demonstrated the robot's adaptivity and performance in multiple scenarios characterized by different requirements in terms of power and workspace. CONCERT has been designed and built by the Humanoid and Human-Centered Mechatronics Laboratory (HHCM) at the Istituto Italiano di Tecnologia in the context of the European Project Horizon 2020 CONCERT.

## 1 | INTRODUCTION

The construction industry is currently facing significant challenges, including labor shortages, high safety risks, and low levels of automation globally Golparvar-Fard et al. (2012), Demirkesen and Tezel (2022), Al-sharef et al. (2024). Although the construction sector plays a significant socio-economic role, providing direct jobs and impacting the global gross product, it is plagued by inefficiency, labor scarcity, quality variation, and a lack of innovation. Robotics has emerged as a transformative technology with the potential to address these issues by enhancing productivity and improving occupational safety. However, construction sites present unique challenges due to their variable tasks and unstructured environments. These sites often suffer from low productivity, an aging workforce, safety concerns, quality inconsistencies, and a lack of innovation Bock (2015). Specifically, they are characterized by:

- Unique workspaces that require different levels of dexterity and payload capacity for various tasks;

- Disorganized environments with scattered materials, necessitating adaptable robotic solutions;
- Tasks that demand significant physical effort, where robots must operate safely alongside human workers.

In tandem with the challenges, there is a critical need to introduce robotic solutions to tackle the occupational issues mentioned before by coordinating effectively with human workers, mimicking the natural collaboration between human partners (Melenbrink et al. (2020)). The next generation of robotic systems, in fact, will need to expand from predictable and isolated industrial scenarios to dynamically changing environments where robots need to work side-to-side with humans. As a result, these robots are expected to be more complex compared to their "traditional" counterparts due to the richness of their sensory system, the variety of tasks they will carry out, and the heterogeneous environment where these tasks will take place.

In response to these challenges, we have developed CONCERT, a new robotic platform focused on providing configurable solutions for unstructured and evolving workspaces. CONCERT addresses the increasing demand for flexible and customizable collaborative robots that align with modern manufacturing trends, such as shorter product lifecycles, increased product variation, and mass customization. These trends require highly configurable robotic solutions that can adapt to less standardized and more dynamic environments, like those found in construction. Further, there is a growing need for collaborative robots with high payload capacities that can ensure safety and efficiency in hazardous, high-effort tasks typical of construction sites while able to be integrated into existing construction sites. Collaborative robots with enhanced physical capabilities and verified safety features can help alleviate these issues by reducing physical strain and hazards for construction workers and improving overall productivity. CONCERT aims to fulfill this need by providing a flexible and adaptive robotic platform that can seamlessly integrate with human workflows, offering tangible benefits in construction environments. The ambition of CONCERT is to create a transition from the current typical cobots to a new concept of all-inclusive configurable robotic platforms, which i) are holistically modular and can be quickly tailored to serve evolving and less structured environments, ii) can regulate online their physical capabilities to execute tasks requiring high forces during coexistence and collaboration with humans and, iii) permit adaptable collaboration, enabling versatile and efficient cooperation with human workers.

In the context of the CONCERT[†] project, a mobile Modular Reconfigurable Robot (MRR) is proposed to perform a variety of tasks such as drilling, painting, and collaborative transportation with just a single platform (Fig. 3). According to the concept presented by the CONCERT project, static robot configurations will therefore be superseded by more dynamic ones where sensors, tools, and even links and joints are subject to in-mission change, depending on the specific task, thus enhancing the flexibility and adaptivity to the numerous on-site scenarios. The control software architecture for MRR should cope with all this variability, and allow a seamless transition (from the user perspective) between different robot configurations.

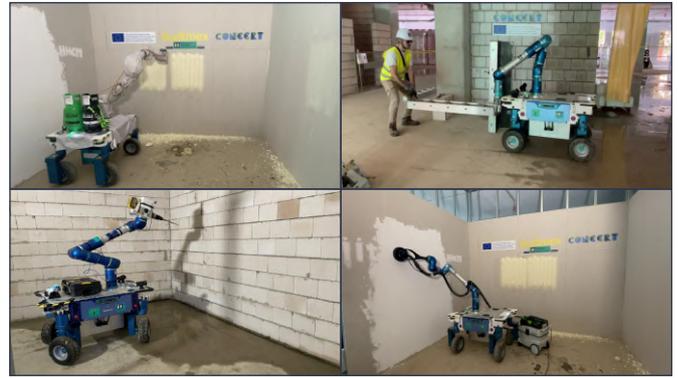

**FIGURE 1** Different morphologies of the CONCERT robot performing different construction tasks.

The rest of this paper will go through previous work from the literature in Section 2. In Section 3 the hardware modules and the specifications of the robot are presented. The method for automatic robot model generation is described in Section 4. Section 5 describes in detail a list of use cases performed with the CONCERT robot. Section 6 shows some experimental results on the proposed use cases that validate the capabilities and the performance of the robot during the execution of on-site tasks.

## 2 | RELATED WORK

### 2.1 | Modular Reconfigurable Robots

MRRs are mechatronics systems composed of modules that can be rearranged to flexibly adapt to changing task requirements. Their versatility, replaceability, and transportability by design render them a promising extension to traditional, fixed-structure robots (Yim et al. (2007), Liu et al. (2016), Guilin Yang (2022)). Contrary to the classic monolithic robots, the modular nature of these systems allows them to be quickly adapted and applied to different applications. Originating in the 1980s (Schmitz et al. (1988), Fukuda and Nakagawa (1988), Benhabib et al. (1989)), MRRs have since then been implemented ins search and rescue (Zhang et al. (2006)), inspection (Wright et al. (2012)), space and exploration operations (Yim et al. (2003)), and recently also in industrial manipulation[‡ §]. Topics receiving particular attention in the last years are automatic controller generation and tuning (Althoff et al. (2019), Nainer and Giusti (2022)) and morphology optimization of such systems (Ha et al. (2018), Zhao et al. (2020)).

---

The primary motivations driving interest in MRRs and inspiring previous research on the topic are as follows. First, the capability to address task changes with minimal down-times of the robot. Second, their cost-effective production and operation: standard modules can be mass-produced and used for a large variety of robots. Third, an improved versatility to realize fit-to-task kinematic arrangements on demand. MRRs can be rapidly customized to meet specific application requirements, in terms of payload, workspace, kinematics, and size. Fourth, the possibility of optimizing the robot structure for a given objective (e.g., cycle time, minimal footprint, energy consumption, or task dexterity). These attributes make MRRs particularly well-suited for scenarios demanding high flexibility where robots must operate in unstructured environments and collaborate with humans. Construction sites exemplify such scenarios, and the limited research on MRRs in construction motivated the work presented in this paper.

## 2.2 | On-Site Construction Robots

Despite the recent advancements in the control of legged platforms, most of the existing robotic solutions for on-site operations in construction sites are skid-steer mobile bases like the ones presented by Gawel et al. (2019), Keating et al. (2017), and Giftthaler et al. (2017). These solutions are characterized by a mobile base, a fixed base manipulator mounted on it, and one or multiple end-effectors that suit a wide variety of tasks, such as drilling, grasping, and even 3D print of buildings. If, on one hand, skid-steer mobile bases offer a stable locomotion on uneven terrain, they suffer from their non-holonomic nature, which is essential when moving in complex and cluttered environments like construction sites. Also, manipulator arms used for these applications are often industrial arms initially designed for fast, precise, and repetitive tasks and cannot handle big payloads and/or are not adaptable to a wide variety of tasks. On the contrary, we designed and scaled the modular units for CONCERT in order to fulfill a range of significant and demanding tasks in terms of contact forces from the interaction with the environment, and maneuverability in confined spaces such as drilling, sanding, and collaborative transportation tasks.

Legged solutions have recently started being explored. Jud et al. (2021) propose an autonomous excavator realized converting an off-the-shelf construction machine into an autonomous robotic system. The robot has been designed with three DoFs hybrid wheeled-legs capable of traversing challenging terrains and with hydraulic actuation for the execution of highly demanding excavation tasks. (Pankert et al. (2022)) propose a hybrid wheeled-legged system able to reconfigure its footprint allows to almost completely remove manipulation induced vibrations. Further, Giftthaler et al. (2017) anticipates an improvement of the existing In-Situ Fabricator (IF), which will be equipped with legs and wheels and will replace the electrical actuation with hydraulic ones. Although CONCERT has not been designed to mount legs, and acknowledging the challenges associated with walking locomotion, its modular design enables the attachment of arbitrarily articulated legs, thereby enhancing the robot's locomotion capabilities.

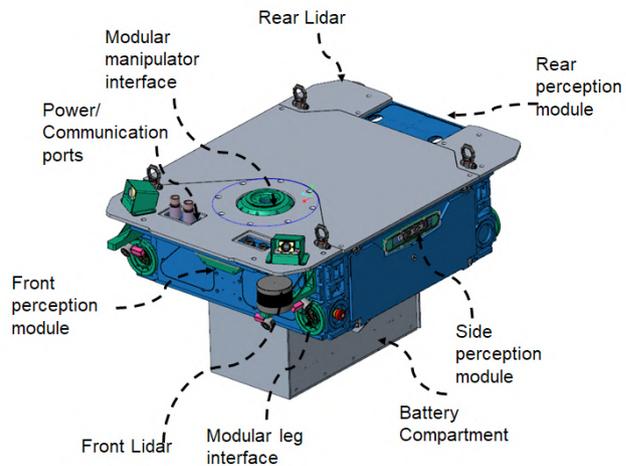

**FIGURE 2** The Mobile Base Hub module with all the onboard sensors and available connectors.

## 3 | SYSTEM OVERVIEW

CONCERT is a modular robotic system designed to be assembled and configured on-the-fly to meet specific user needs. It consists of multiple modules—both actuated and passive—that can be added or omitted without affecting the functionality of the entire system. *Joint* modules of various kinematics and sizes are used to add degrees of freedom to the kinematic chain, *Hub* modules enable branched kinematics, *Link* modules serve as passive extensions, and *End-Effector* modules complete a kinematic chain with the required tool for a given task. This modularity supports the creation of both robots functioning as mobile robots and as fixed-base manipulators. The mobile base of CONCERT is also following a modular implementation concept. It consists of a *Mobile Base Hub* module and four *Steering + Wheel* modules, which will be introduced in Section 3.1. This design choice allows the system to adapt for indoor or outdoor mobility by interchanging Wheel modules.

As a note on notation, we will use capital letters for the names referring to the modules (e.g., Joint module) to differentiate them from the general concepts they represent (e.g., a joint as a connection between bodies that allows movement).

Given the focus of this paper on mobile construction robots, we assume the presence of a mobile base in all configurations, making the Mobile Base Hub an essential component and the foundational module of any CONCERT robot instance. Due to its critical importance in the system, we will describe the characteristics of the Mobile Base Hub before delving into the details of the other modules. The Mobile Base Hub provides connection interfaces for four legs and one arm, as illustrated in Figure 2. It also houses the robot's computational and power units, along with the majority of its sensors. Specifically, it includes the components listed below.

**Embedded computers.** Inside the CONCERT Mobile Base Hub, three computational units are incorporated to support the computational needs of the platform from the control, perception, and interaction perspectives. These computational units are:

- an RT Control PC: a SoC (System on Chip) responsible for the Real-Time control of the EtherCAT network of the CONCERT actuators;
- a Navigation PC: general purpose PC with a powerful GPU, responsible for the navigation tasks of the CONCERT mobile base;
- a Manipulation PC: general purpose PC with a powerful GPU, responsible for the manipulation tasks of the CONCERT configurable arm;

**Perception system.** The above-mentioned computational units use a set of sensors installed on the Mobile Base Hub to accomplish their mission:

- On the front left and front right parts of the mobile base, there are two DALSA Genie Nano M2590 cameras for Human 3D pose detection and estimation;
- In the front and rear parts of the mobile base, there are two sets of sensors composed of a depth camera (Intel RealSense D435i), a LiDAR (Velodyne Puck VLP-16), and two ultrasound distance sensors. These are used for both collision-free navigation and manipulation purposes;
- On the left and right sides of the mobile base, there are two copies of the same set of sensors composed of an Intel RealSense T265, which is a tracking camera, and two ultrasound distance sensors. These sensors are used for SLAM and collision-free navigation purposes;

**System power.** The system supports both tethered and battery-powered operation. The battery requires 3 hours to fully charge and with a capacity of $72\,$Ah can provide 6 to 8 hours of autonomy, depending on the activity and usage of the CONCERT platform, considering a nominal consumption of $9\,$A;

**Safety devices.** Four physical safety buttons are installed at the corners of the mobile base, allowing for immediate shutdown of robot operations in case of an emergency. Additionally, a wireless safety button receiver is mounted on the top plate of the base to safely stop the robot remotely.

## 3.1 | Modular Units and End-Effectors

Each connector on the Mobile Base Hub module features an electromechanical interface (EMI) that enables power, data, and mechanical connections to be made with the corresponding connector. Consequently, any module compatible with the CONCERT platform will include such an EMI on both its input and output connectors, allowing for modules to be interconnected. The available catalog of robotic modules for use with the CONCERT platform is shown in Figure 3 , which also includes the corresponding names for each module. The set of modules includes:

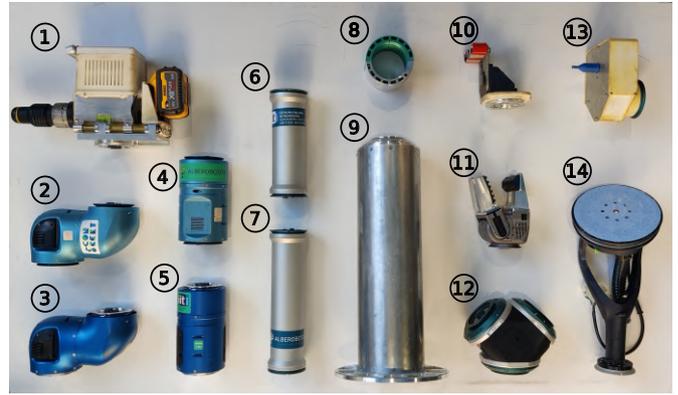

**FIGURE 3**  Tabletop view of all components needed for assembling the reconfigurable robot manipulator: (1) Drill module, (2) Elbow A Joint, (3) Elbow B Joint, (4) Straight A Joint, (5) Straight B Joint, (6) Passive Link $0.3\,$m, (7) Passive Link $0.4\,$m, (8) clamps for modules interconnection, (9) Passive Base Link, (10) Passive Gripper, (11) Active Gripper, (12) Torso Hub, (13) Spraying Tool module, and (14) Sanding Tool module.

**Elbow Joint modules.** They enable rotation in a direction perpendicular to the input EMI axis (pitch direction). These joints are available in two sizes, A and B, based on the actuators they use, which are categorized as *Large A* and *Large B* (see Table 1). In an anthropomorphic manipulator, the higher-torque *Elbow A Joints* can be used as shoulders or elbows, while the lower-torque *Elbow B Joints* as elbows or wrists.

**Straight Joint modules.** They introduce a rotation in a direction parallel to the input EMI axis (yaw direction). Similarly to the Elbow Joints, they also come in two sizes, A and B, which are selected depending on the torque requirements of the robot.

**Steering + Wheel modules.** They are connected exclusively to the Mobile Base Hub leg interfaces and embed both a *Steering Joint* and a *Wheel* module. The Steering joint uses a *Medium* actuator, while the Wheel joint a *Small B*, both faster and lower-torque actuators compared to the ones used in the arm (see Table 1).

**Passive Link modules.** They are passive non-actuated extensions that permit to adjust the size of the manipulation systenm and its workspace eventually by incorporating then between the active Elbow and Straight joint modules. The modules that provide straight displacement are available in two sizes: $0.3\,$m and $0.4\,$m, while those that introduce an angular offset come in two variants: $90°$ and $45°$. A passive link designed as the first connection after the Mobile Base Hub is also available. This module, called *Passive Base Link*, is used when the manipulation subsystem needs to operate at higher heights in the range of $2.4 - 2.9m$, raising the base of the manipulation subsystem by $0.6\,$m.

**Hub modules.** They allow for the branching of multiple chains, facilitating the construction of tree-like robots. In addition to the Mobile Base Hub, the *Torso Hub* is also available, enabling bifurcation into two branches, thus making it possible to create humanoid-like upper bodies with a pair of arms.

| Type | Gear ratio | Joints | Maximum velocity [rad s⁻¹] | Torque peak - continuous [N m] | Power peak - continuous [W] | Mass [kg] |
|------|-----------|--------|------------------|-----------------|----------------|-----------|
| Large A | 160 | Elbow, Straight | 2.14 | 460 - 160 | 930 - 326 | 2.5 |
| Large B | 120 | Elbow, Straight | 2.85 | 314 - 120 | 930 - 326 | 2.0 |
| Medium | 120 | Steering | 8.1 | 127 - 34 | 820 - 259 | 1.3 |
| Small A | 100 | Gripper | 11.7 | 55 - 17 | 556 - 179 | 1.0 |
| Small B | 29 | Wheel | 13.7 | 24 - 17.5 | 337 - 244 | 2.7 |

**TABLE 1**  Actuators specifications.

Together with the set of basic modules to compose the robot kinematics, a set of specialized end-effectors, each designed for a specific task, are also available:

**Drill module.** Designed by repurposing a commercial hammer drill, with the handle removed and the battery repositioned. Additionally, an Intel RealSense D435i depth camera has been integrated to provide perception data for the drilling operation, and the standard EMI has been integrated at the input connector.

**Spraying Tool module.** Its main functionality is to control the mixing of the two pressurized liquid components of the insulation material used in the spraying task. To achieve this, two electrovalves are used to control the flow of the two liquid components. The components, exiting from the outlets of the electrovalves, are directed to a passive, spiral-shaped blender, which mixes the materials before they exit through the blender's outlet.

**Passive Gripper modules.** Designed by adapting professional passive tools for handling panels and tubes. These modules are passive self-locking grippers, which are intended to be used in collaboration with the operator, who works alongside the robot. Unlike the other active modules, they do not feature an EMI, but only the mechanical mechanical interface.

**Active Gripper modules.** An active version of the passive grippers for panels and tubes, designed with a modular jaw-type gripper (Del Bianco et al. (2024)). The bottom jaw is fixed, while the top jaw is powered by a single actuator of type *Small A* (see Table 1) within the gripper's main body. Both jaws are modular, allowing for easy interchange or replacement with different-shaped jaws to accommodate various tools or object sizes. Dedicated jaws have been designed for gripping panels and tubes similarly to the passive version but autonomously.

**Sanding Tool module.** A commercial sander was modified to integrate the standard modular connection intergace permitting its integration with the CONCERT manipulation system. The sander module is powered externally, requiring an external power cable and a dust collection tube.

## 3.2 | System specifications

The overall system possesses the following specs:

**Maneuverability:** the robot can move omnidirectionally and perform tight turning maneuvers.

**Footprint:** to ensure mobility through narrow width passages (i.e., corridors and doors), the footprint is limited in width up to $0.85$ m and in length up to $1.30$ m.

**Terrain traversing:** the platform can move on uneven ground overcoming $0.1$ m high debris and traversing slopes up to 10 °.

**Carrying payload capacity:** the platform can carry and transport tools and material payload of $200$ kg

**Stability under interaction forces:** the floating base mobile platform can maintain adequate stability margins for its center of pressure while the carried manipulation system exerts forces up to $200$ N to the environment during demanding tasks such as drilling.

**Speed range:** the mobility system is capable of regulating its speed up to the typical walking speed of a human in the range $6$ km h⁻¹ to $8$ km h⁻¹ on flat ground.

**Payload:** the system's modularity enables a variable payload capacity. A standard 6-DOF robot, using the currently available modules, supports a nominal payload of 10 kg to 25 kg. However, optimizing the robot's trajectory and morphology for a specific task can allow it to handle heavier payloads.

**Workspace:** with the available modules, the robot's vertical workspace ranges from $0.0$ m to $3.0$ m, leveraging the available passive linkages.

# 4 | AUTOMATIC TOPOLOGY AND MODEL RECONSTRUCTION

To ensure the practicality of CONCERT's adaptability to a wide range of tasks, a method for rapidly and automatically detecting and generating the robot model is needed. To this end, the method proposed in Romiti et al. (2022) for reconstructing the topology of reconfigurable EtherCAT-based robots was exploited in the CONCERT robot model generation procedure. This approach enables the robot to automatically identify its morphology, which can change based on the combination of body modules, each operating as an EtherCAT slave. The kinematic and dynamic models of the robot are obtained in an automatic way and stored in URDF format as soon as the physical robot is assembled or reconfigured. The proposed method can support arbitrary serial kinematic tree-like configurations. Finally, it permits rapid and automatic adaptation of the robot by reconfiguring all its software components on-the-fly, almost without any intervention by the user. In this section, we'll recap the functioning principles of the method and show some examples of the robot model generation for the CONCERT robot.

## 4.1 | Reconfiguration Process Overview

Initially, the user manually assembles the robot by connecting the modules to realize the desired kinematic configuration suitable to the requirements of the targeted automation task. This is the only manual step where the user input is required. As soon as this step is completed and the physical robot is assembled from the interconnected modules, a network communication is established among the modules, and the network topology can be discovered, as described in Section 4.2. This enables to reconstruct the parent/child relationship among the modules. Each type of module also contains a unique identifier that permits after the discovery of the network and the identification of the interconnected modules, to access a database and obtain all parameters of each module. This information is then used to reconstruct the physical properties of the robot: kinematic, dynamic model, and semantic information such as which module is an end-effector, or which modules are part of the same kinematic chain. This will be described in Section 4.3. Finally, the software architecture is dynamically reconfigured at the end of the robot reconfiguration permitting the user to operate the robot immediately and perform the required task right after the completion of this process. This will be more thoroughly discussed in Section 5.

The rapid reconfigurability of the robot allows the user to quickly iterate the entire process to arrive at the most effective solution or to quickly adapt the robot configuration to the new task and/or workspace settings. The same process can be used in the design phase where characteristics of different robot arrangements are evaluated in a simulation environment and iteratively new designs are virtually composed, for example with the goal of minimizing a certain objective function as evaluated in Romiti et al. (2021 2023), Külz and Althoff (2024).

## 4.2 | Network Topology Recognition

The CONCERT robot uses EtherCAT as communication network technology, which is particularly suitable for tree-like robots because allows the daisy chaining of nodes and supports multiple kinds of network topologies (line, tree, or star). The CONCERT robot is composed of a number of body modules, each of them incorporating an EtherCAT slave controller (ESC). By interconnecting these modules to realize a reconfigurable robot, an EtherCAT network is formed with a centralized master and the modules being the slaves. We leverage on the formed EtherCAT network and protocol to retrieve the necessary information about the slaves and the network topology. We refer the reader to Appendix A for the functioning details of EtherCAT networks and the algorithm for network topology reconstruction presented in Romiti et al. (2022).

To summarise, any physical connector of a module is associated with a port of the ESC allowing to establish a 1:1 correspondence between the physical topology of the robot and the network topology. The result of this step is a graph structure $\chi$ representing the network topology with the slaves being the nodes and the connection among the chips' ports being the edges of the graph.

As described in Appendix A, the EtherCAT protocol restrictions force a module's upstream port and its associated connector to be static, preventing a module to be connected upside down and therefore restricting the possible connections between two modules. This is signaled to the user by labeling the connector associated with the upstream port as "input connector", to avoid errors during the assembly operations that would result in an incorrect network topology. In order to allow for mounting a module upside down, a hardware circuit that can swap the "input connector" on demand as in Yun et al. (2020) would be required, but was not implemented on CONCERT modules, as it would not have increased much the set of possible robot solutions.

## 4.3 | Robot Physical Topology Recognition

At this point, we identified the graph $\chi$ describing the network topology with each node representing an EtherCAT slave (Section 4.2). For each node, we can extract the information fully characterizing the module hosting the slave from the centralized database (Section 4.3.1). The convention for frame assignment described in Section 4.3.2 allows us to expand each slave node into multiple physical nodes as described in Section 4.3.3. This information is now aggregated into a tree-like data structure $\phi$, where each node represents a physical body module. This new graph $\phi$ contains bodies as nodes, and joints and physical connections as edges; rather than slaves and ports. This tree reflects the physical topology of the robot and can be converted effectively to URDF (Unified Robot Description Format), which will be exchanged among the different software agents. With the given robot model we can compute kinematics and dynamics quantities as according to Featherstone (2008).

### 4.3.1 | Module Database

After creating $\chi$, for each slave we have a *Module identifier*, which serves as a key to look up the module properties in a centralized module database. To each *Module identifier* corresponds a *Module description* file that stores all the properties relative to that kind of module inside the centralized module database. These properties comprehend at least:

- The module type: Joint, Link, Hub or End-Effector.
- The list of connectors associated with the module, their type (input/output), and other characteristics (e.g. size of the connector).
- The coordinate transformations between the coordinate frames assigned to the module.
- The inertial parameters of the module. This includes the center of mass coordinates w.r.t. the body frame, the module mass along with the module inertia parameters w.r.t. the center of mass. For

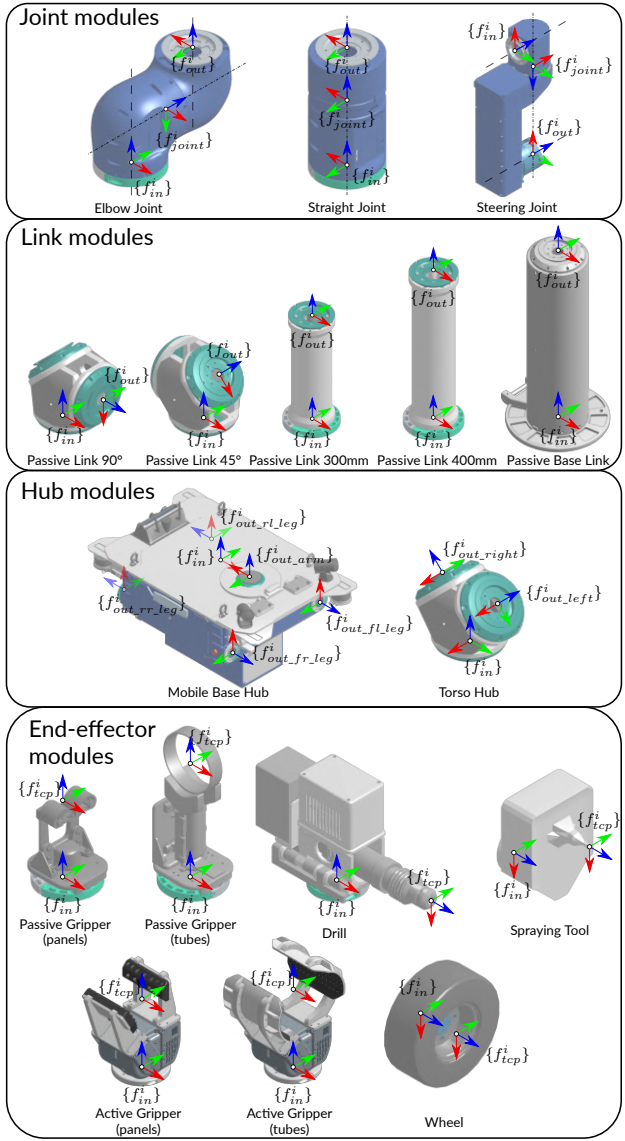

**Joint modules**

Elbow Joint    Straight Joint    Steering Joint

**Link modules**

Passive Link 90°   Passive Link 45°   Passive Link 300mm   Passive Link 400mm   Passive Base Link

**Hub modules**

Mobile Base Hub    Torso Hub

**End-effector modules**

Passive Gripper (panels)   Passive Gripper (tubes)   Drill   Spraying Tool

Active Gripper (panels)   Active Gripper (tubes)   Wheel

**FIGURE 4**  Example of frames definition for the different types of modules and their connections. The circles indicate frame origins. The XYZ coordinates correspond to the RGB color channels, where X aligns with Red, Y with Green, and Z with Blue

JSON objects are used to store information about each module. This information includes details about the module's kinematic, dynamic, and geometric properties. The same format used in Mayer et al. (2024) is used to store the data. The JSON database of CONCERT modules is publicly available in the *concert_resources*[¶] repository.

### 4.3.2 | Module Coordinate Frame Assignment

We introduce a coordinate frame convention to formalize the automatic model generation. We place a separate coordinate frame in the center of each EMI as well as on each actuated joint axis. We'll call "input frame" the one centered in the EMI associated to the "input connector", "joint frames" the ones centered in the joint axis, and "output frames" the ones centered in the EMIs associated to the "output connectors".

These frames are sufficient to define all the kinematic properties of a module. An additional reference frame on each moving body center of mass (CoM), with axes parallel to the body frame (either the input frame or a joint frame), locates the inertial properties of the module.

These coordinate frames {f} and the relative transformations **T** between them allow to unequivocally derive the robot physical model (kinematics and dynamics). The respective geometric and inertial module parameters are stored in the centralized module database of Section 4.3.1 and can be collected in the data structure $\phi$, which can then be translated in URDF (see Section 4.3.3). The URDF represents the robot as a series of links with certain inertial parameters connected by either fixed, prismatic, or revolute joints. Subsequently, other software modules parse the URDF to generate the dynamic model.

For what regards notation, in Romiti et al. (2022) is presented a naming convention for the frame axes that can be assigned directly after retrieving info from the EtherCAT network. To make the notation lighter and more readable, we'll use a simplified version of it here. The trailing superscript is the index associated with the module, while the trailing subscript refers to the port of the type of frame it is (input, output, joint, etc.): $\{f_{type}^{module\_idx}\}$. As a convention in this paper, the $Z$ axis of the module interface is perpendicular to the plane defined by the EMI and always points "downstream": away from the origin of the input frame. The coordinate frames are kept parallel to the input frame where possible. If a reorientation of one coordinate frame with respect to the input frame is necessary, the approach is to rotate about the minimum number of axes. The *module_idx* index is associated to the module depending on the order in which the modules are discovered in the network. This is not exactly coincident to the position of the slave in the EtherCAT ring, as some modules might embed more than one ESC.

Figure 4 shows the complete library of modules available at the time of writing the paper, and how the coordinate frames are defined for: Joints, Links, Hubs and End-Effectors modules. Links and joints have just one input and one output connector, so their input and output frames are named $\{f_{in}\}$ and $\{f_{out}^i\}$ respectively. $i$ in this case represents the

Joint modules, the parameters are stored separately between the upstream body and the downstream body of the joint.

- Semantic information describing the common purpose of the module, which identifies for example an End-effector module as a 'gripper' or a 'foot' or a 'wheel'.
- Parameters for possible kinematic, differential, or dynamic constraints associated with the module. Joint motion range, torque, and speed limits are common constraints.
- Links to a 3D mesh file graphically representing the module body, and coordinate transformation between the upstream frame of the body and the mesh origin.

*module_idx*. Joints also have an additional frame defined at the joint axis of action, named $\{f_{joint}^i\}$. The $Z$ axis of this frame is always parallel to this axis. Hub modules instead have multiple output connectors. The number of available output connectors is determined by the number of ESCs integrated in the module. In the case of the Torso Hub module, which has just one ESC integrated, there are two output connectors (it could have a maximum of three). In the case of the Mobile Base Hub module, two ESCs are integrated, so it provides five connectors where a branch can be started. Connector names are used to distinguish between them and are stored in the database together with the transformation matrices between the associated frames and the input frame. It is visible that in any case, for each input or output connector, there is a coordinate frame centered with the associated EMI. End-Effector modules have one input frame $\{f_{in}^i\}$ and no output frames since they don't allow any further connection and are used to close a kinematic chain. Depending on the type of end-effector they have an additional frame defined either at: (i) the Tool Center Point (TCP) $\{f_{tcp}^i\}$, for end-effector meant to be mounted on arm chains, (ii) the wheel center $\{f_{wheel}^i\}$, for end-effector meant to be mounted on leg chains. Certain End-Effectors such as the Wheel module are actuated by a joint and therefore, like Joint modules, have an additional frame $\{f_{joint}^i\}$ defined at the joint axis of action.

The CoM coordinate frame, as per convention, has the same orientation as the input frame and origin at the CoM of the body. The inertia tensor stored in the database will be referred to this frame. For Joint modules, as these are composed of two moving bodies, we have two frames one for the upstream (proximal) body and one for the downstream (distal) body.

When connecting two modules through two of their EMIs, the frames associated to them will by default be coincident, unless they have an orientation offset about the common axis. The possibility of allowing an orientation offset would mean having a more complex EMI, since it must allow both the mounting and the detection of said offset, and was not realized for the CONCERT modules, although is potentially possible.

Figure 5 shows an example of robot morphology with the purpose of showing the frame convention when modules are assembled. The Mobile Base Hub is the first module to be discovered in the network and is therefore assigned the index 0. The four leg connectors of the Mobile Base Hub are distinguished by their positioning relative to the input frame: front-rear and right-left. To each of these connector, a Steering Joint module is connected and a Wheel module after that. The last connector of the Mobile Base Hub is used to connect the arm chain, which is composed by a series of Elbow and Straight Joint modules, Passive Link modules, and a Drill End-Effector module to close the chain.

It is interesting to notice how any output frame of a module is coincident with the input frame of the next module, when these two modules are connected together. The reader can refer to Figure 4 for reference. Since there cannot be an orientation offset between the frames of two coupled EMI, the information obtained in Section 4.2 is sufficient to obtain the full model by using only the data stored in the database. In fact, since we have stored in the database the transformations $T_{in,joint}$ and $T_{joint,out}$ for Joints, the transformations $T_{in,out}$ for Links and Hubs, and the transformations $T_{in,tcp/wheel}$ for End-effectors, obtaining the kinematic model of a complex robot like the one in the figure becomes trivial. More details will be given in the next subsection.

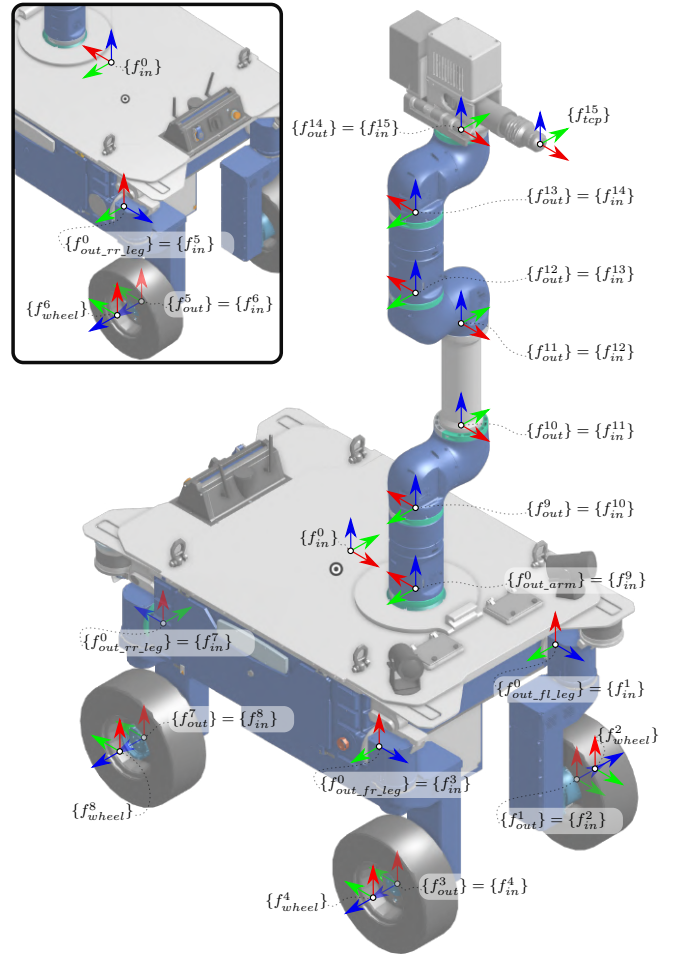

**FIGURE 5** Example of frame placement when connections between modules are established for different types of modules. In the top left tile, a view from the rear shows the frames also for the modules whose view is obstructed in the main image.

### 4.3.3 | Modelling and URDF/SRDF Generation

The coordinate frames convention introduced in Section 4.3.2, permits to unequivocally derive the robot physical model and store it in the graph $\phi$. We can then automatically convert this data in URDF, which serves as input to the other software modules. Since reference frames associated with paired connection interfaces will be coincident, the relative kinematics between two subsequent modules will depend solely on the parameters of the parent module. Following the notation in Featherstone (2008) for kinematic trees, we can denote as $\lambda(k)$ the parent of a module of index $k$. The $T_{\lambda(k),\,k}$ transformation matrix between modules $\lambda(k)$ and $k$, can be defined as the one between the input frames of the modules $\{f_{in}^{\lambda(k)}\}$ and $\{f_{in}^k\}$, by

$$T_{\lambda(k),\,k} = T_{f_{in}^{\lambda(k)},\,f_{in}^k} = T_{f_{in}^{\lambda(k)},\,f_{out}^{\lambda(k)}} \cdot \underbrace{T_{f_{out}^{\lambda(k)},\,f_{in}^k}}_{=1} = T_{in,out}^{\lambda(k)} \qquad (1)$$

where $T_{in,out}^{\lambda(k)} \in SE(3)$. In case the module has multiple outputs here we consider as output frame one of the many defined in the *Module description* entry in the database.

For a module $k$, which can be any node in $\chi$, the transformation depends only on the parent node $\lambda(k)$ and the edge connecting the two nodes. As mentioned in Section 4.2 the edges of $\chi$ store which port of the ESC associated with $\lambda(k)$ is used to connect with the one associated with the module $k$, that from a physical topology point of view, identifies which connector of $\lambda(k)$ is used. Therefore, the relative forward kinematics between two modules $a$ and $b$ can be computed by traversing the graph from node $a$ and iteratively calling (1) until $b$ is reached. The modular kinematics for any module $k$ is:

$$T_{in,out}^k = \begin{cases} T_{in,joint}^k \cdot e^{\hat{s}_j^k q_k} \cdot T_{joint,out}^k, & \text{if } type = \text{Joint} \\ T_{in,out}^k, & \text{if } type = \text{Link, Hub} \\ T_{in,tcp}^k, & \text{if } type = \text{End-Effector} \end{cases} \qquad (2)$$

where the above transformation matrices are as follows: $T_{in,joint}^k$ from frame $\{f_{in}^k\}$ to $\{f_{joint}^k\}$, $T_{joint,out}^k$ from $\{f_{joint}^k\}$ to $\{f_{out}^k\}$, describing the *proximal* and *distal* parts of the Joint module respectively, $T_{in,out}^k$ between input port and the output port for Link/Hub modules, and $T_{in,tcp}^k$ between input port and the TCP of End-Effector modules. $q_k$ is the joint displacement of module $k$. $\hat{s}_j^k \in se(3)$ is the twist of the joint of module $k$ expressed in frame $\{f_{joint}^k\}$. The 6-D coordinate vector $s_j^k$ representing the twist coordinates of the joint axis is constant, with $s_j^k = [0,0,0,0,0,1]^T$ for revolute joints and $s_j^k = [0,0,1,0,0,0]^T$ for prismatic joints.

By traversing the graph $\chi$, each node can be expanded by applying (2) and using the data retrieved from the database, to obtain the graph $\phi$. For example, a Joint module node is expanded into two nodes representing the proximal and distal bodies, connected by an edge representing the actuated joint. The nodes store the dynamical parameters of each moving body and the edges the transformations relating the bodies to each other.

In Figure 6 we show an example of the graphs $\chi$ and $\phi$, and a snippet of the corresponding URDF, for the robot in Fig. 5. In Figure 6.a a simplified version of the network topology graph $\chi$ is shown, where each node (ESC) is labeled with the name assigned to the corresponding module for URDF generation to make the graph more readable. The *Module identifier* stored in each ESC encodes which type of module is representing. For example, we know that ESC 1 and 2 are part of a Mobile Base Hub module, and ESC 3 is part of a Joint module. The nodes are colored according to the corresponding module type, and the edges are colored in grey. The edges represent the connections between the ports of the ESCs and, therefore, determine the parent/child relationship between each ESC.

In Figure 6.b a visual representation of the relationship between the network topology and the real robot morphology is shown.

In Figure 6.c the physical topology graph $\phi$, is shown, where each node represents a physical body module and each edge represents a physical connection between two bodies. The graph $\phi$ is obtained by expanding the nodes of the graph $\chi$ according to the information stored in the database, in order to switch to a graph that holds the physical properties of the robot:

- For Joint modules we'll expand each node in two nodes representing physical bodies, connected by one vertex representing the transformation between the two reference frames associated to the two bodies: $T_{in,joint}^k \cdot e^{\hat{s}_j^k q_k}$.
- The static connections established by two mating EMIs are also represented as vertexes. In the figure, the vertexes corresponding to static connections are colored in grey. When two Joints are connected for example, a vertex is created to connect the *distal* part of the upstream Joint module and the *proximal* part of the downstream Joint module. This vertex represents the static link between these two bodies which is established mechanically and the rigid transformation $T_{joint,out}^k$ of the parent module. In a similar way vertexes are created also when other types of modules are connected. For Link and Hub modules, the transformations represented by the vertex is $T_{in,out}^k$, which is stored in the database. End-effector modules do not have available output connectors for further connections, so no vertex can be created from them.
- Hub and Link modules are expanded in a single node representing the physical body of the module. As mentioned before, Hubs can embed more than one ESC, which is the case here for the Mobile Base Hub module, comprising two ESCs. Although in the $\chi$ graph Slave 1 and 2 are two different nodes, when passing on to $\phi$ these two become one single node. In fact from the module identifier of the first two slaves we know they are part of a single module, and the physical body representing the Hub module is a single one.
- End-Effector modules that are actuated by a joint get expanded in the same way as Joint modules, while non-actuated one get expanded as Link modules, although an extra link is added with null mass and a vertex to indicate where the TCP frame is.

Note that in Fig. 6 the colors in graph $\chi$ are used to show in what nodes and edges the slave expands to in graph $\phi$. The names of nodes and edges do not refer anymore to the network topology or the slave position in the EtherCAT ring, but instead to the physical structure of the robot. For naming bodies and joints, we'll use a *joint index* that starts from 0 and increments only after a moving joint (i.e. a Joint module or an actuated End-Effector module such as Active Gripper or Wheel modules) has been added. A second index is used when passive links (i.e. Link or Hub modules) are present: the *link index* starts from 0 and increments for each passive link added after the last moving joint. The *link index* resets to 0 when a moving joint is added. Moreover, no more references to the ESC ports are left in $\phi$, instead, some tags (A, B, C,...) are appended to the node name to specify to which *kinematic chain* they belong. A new chain can be started only from a Hub module, while Joint and Link modules can just extend the previously created chain. End-effector modules close the chain as they do not have output connectors.

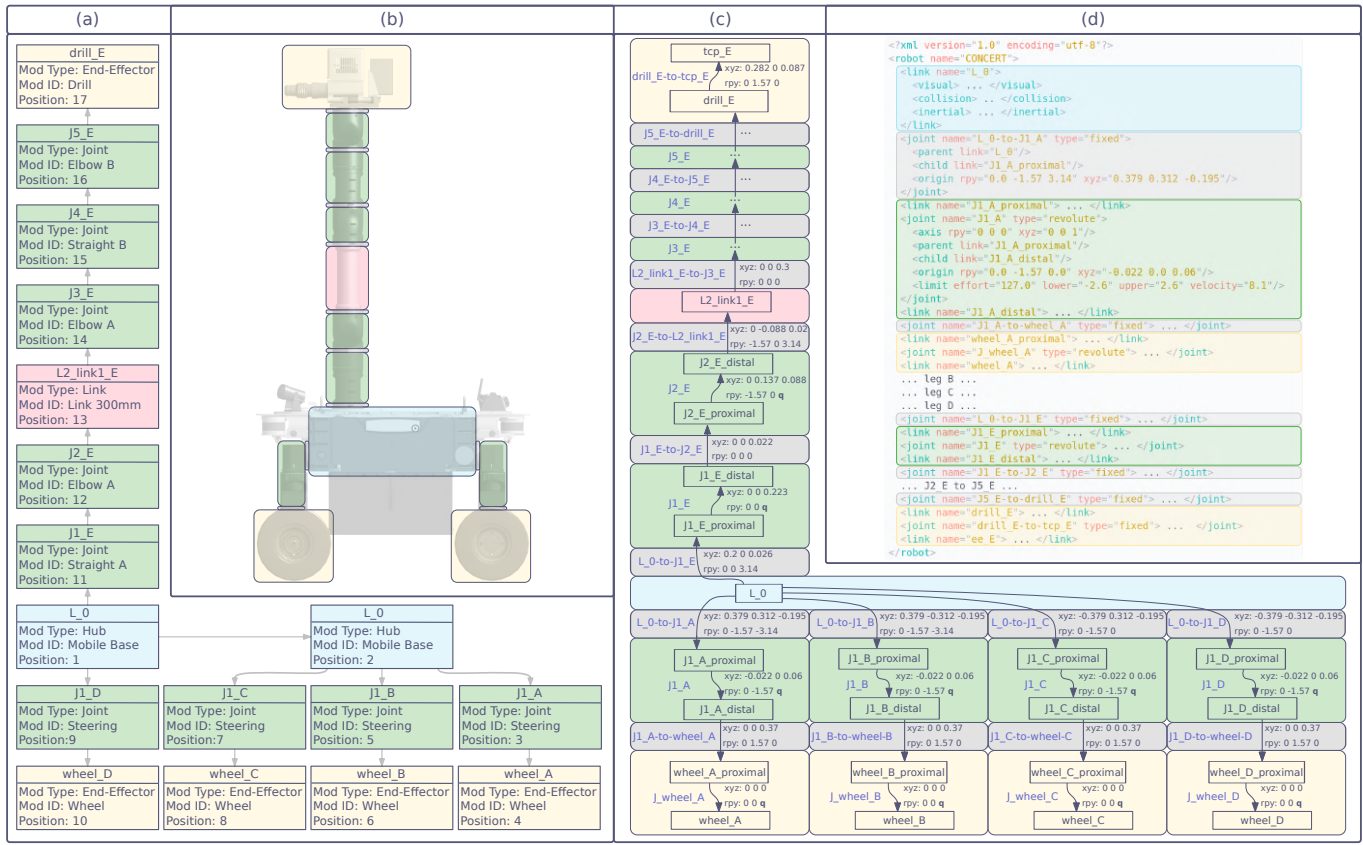

**FIGURE 6** Example of robot physical topology reconstruction for the robot in Figure 5. The network topology graph is shown first, then expanded in the physical topology graph, and lastly, a snippet of the generated URDF is provided. The different phases are depicted in the picture with colors propagated throughout them to indicate how the different modules are processed. Different colors are used for each module type: light blue for Hub modules, green for Joint modules, pink for Link modules, and yellow for End-Effector modules. The static connections between neighbor modules are colored in grey.

Considering the URDF format (an XML format) as a de-facto standard for ROS-based libraries, we convert the graph $\phi$ to a URDF file to describe the kinematics and dynamics model of the robot, representing the robot as a series of *link* elements (which consists of inertial, visual and collision properties) connected by either fixed, prismatic or revolute *joint* elements. The mapping between nodes and edges of $\phi$ and the URDF XML elements is therefore 1:1. In Figure 6.d a snippet of the URDF file obtained from $\phi$ is shown. Only certain nodes are shown in the URDF for brevity. Note that the static physical connection between two *links* (for instance between the distal and proximal body of two successive Joint modules) is represented by a *fixed joint* element, leaving to the URDF parser of the dynamic library of choice, the task of computing the dynamic parameters of the composite body. The advantage of this format is mainly the easiness of exchange among different software modules, and the large number of libraries that can take this as input for the robot model.

Semantic information about the robot, such as a description of the end-effectors, kinematic chains, and joints composing them, is instead written in a SRDF file (Semantic Robot Description Format), initially introduced by the MoveIt! framework Coleman et al. (2014) to complement the URDF. In our software architecture, the SRDF serves to configure the API to control the robot, providing methods for the whole kinematic chain, single joints, or end-effector. For the example in Figure 6, the SRDF file would contain information on the five chains that compose the robot. Four chains are *leg* chains, as they contain a Wheel End-Effector module at the end, and one chain is the *arm* chain, as it contains a Drill End-Effector module at the end.

The URDF can be used as input for many libraries for path planning, collision avoidance, control, but in particular for kinematic and dynamic computations. The dynamic library of choice derives unequivocally the kinematic and dynamic model from the URDF. Particularly efficient implementations of dynamic libraries, suitable for computations in a high-frequency real-time control loop, are the ones using spatial algebra notation Featherstone (2008) such as KDL[#], Pinocchio Carpentier et al.

---

[#] https://www.orocos.org/kdl.html

(2019) or RBDL Felis (2017). In our current implementation the latter was used, although any of these libraries allows to numerically compute from the URDF input all the main kinematic and dynamic quantities.

## 4.4 | Reconfiguration Process Benchmark

The topology recognition and model reconstruction procedure presented in this section has been tested on several occasions in the field. The whole procedure, from network topology recognition to URDF generation, has been implemented inside the *modular*[||] package, which provides also a web-based GUI where the model reconstruction process can be started, the generated URDF visualized, and a ROS package with the description of the "discovered" robot deployed on the embedded PC. Such ROS package will contain URDF, SRDF, launch files, and XBot2 configuration files (more info will be given in Section 5.1) that allow starting the robot right after this reconfiguration step. Having a web-based GUI for performing this is crucial for a worker in the field as the GUI can run on any handheld device such as tablets or smartphones.

From the GUI it will also be possible to customize some details of the robot model that cannot be discovered from the network scan. In general, all modules can have some add-ons that can be manually added to the model, since are not connected to the EtherCAT network. These can be, for instance, the tools used in end-effectors that allow quick tool change (like the commercial drill used for our Drill end-effector module) or additional sensors that can be present or not (like cameras, sonars, etc.).

In Figure 7 are shown three examples of the reconfiguration process for different robot morphologies in different scenarios. For each run, is shown: first the manual assembly of the modular robot performed by the users (phase a), the start of the model reconstruction procedure by interacting with the GUI (phase b), the end of the procedure with the reconstructed model of the robot, when the user can perform some manual customization of the homing configuration and of any add-ons he/she wants to use for the next task (phase c), and finally the robot performing the desired task after the robot model has been generated (phase d). More details on how the software stack allows to run the tasks shown in the figure (drilling, collaborative transportation, etc.), based on the inputs produced in this phase will be given in the next section.

To validate the proposed procedure in terms of reconfiguration time, we have measured the time needed in the several reconfiguration experiments carried out in the field. The results of such experiments are summarized in the statistics of Figure 8. The average re-configuration time comes out to be approximately $\tilde{T}_{\text{reconf}} \approx 70$s per module, of which $\tilde{T}_{\text{mount}} \approx 40$s seconds for mounting and $\tilde{T}_{\text{unmount}} \approx 30$s seconds for un-mounting a module. The time needed by the software components from the point when the physical assembly is finished to have the robot operational was also measured in these trials and results as approximately $\tilde{T}_{\text{software}} = \tilde{T}_{\text{URDF}} + \tilde{T}_{\text{custom}} + \tilde{T}_{\text{startup}} \approx 60$s. This time comprises the time needed by the topology reconstruction and model reconstruction procedure to generate the URDF/SRDF ($\tilde{T}_{\text{URDF}} \approx 5$s), the time needed by the user to perform the customization of homing and add-ons ($\tilde{T}_{\text{custom}} \approx 30$s), and the time to start the EtherCAT master, XBot2, and initialize all low-level controllers ($\tilde{T}_{\text{startup}} \approx 25$s).

Reconfiguring a robot composed of 9 modules, like the 6-DoF arm with two passive links and one drill end-effector in experiment 1 of Figure 7, to another arm still composed of 9 modules, takes then $\tilde{T}_{\text{tot}} = \tilde{T}_{\text{reconf}} \times 9 + \tilde{T}_{\text{software}} \approx 680$s, so slightly more than eleven minutes. It's worth noticing that in this case, the time needed from the software side is almost negligible with respect to the one needed for the physical reconfiguration from the hardware side, which is about one order of magnitude higher and it's clearly the bottleneck for a faster reconfiguration. Anyhow, considering that this reconfiguration could allow us to completely switch task, we believe this gives an indication of the flexibility and ease-of-reconfiguration such a platform can provide. To give an idea, the reconfiguration could mean we have added/removed DoFs, changed the end-effector, and modified the robot workspace, for example switching from a drilling robot thought to operate at 3m height to a plastering robot thought to operate at 1m height.

## 5 | THE CONCERT SOFTWARE ARCHITECTURE

The large variability of tasks usually performed by a human worker in a construction site should reflect in a large configurability of those robotic solutions aiming to aid the worker in the field. In the previous sections, we showed how MRRs are an approach to cope with such variability from the hardware side. Simultaneously, the control software architecture for these types of robots is expected to support and ease, as much as possible, the development of control algorithms that can take into account all this variability.

In this section, we'll present how exploiting topology recognition and automatic URDF generation capabilities (detailed in Section 4), the software stack can dynamically reconfigure and enable a configuration-agnostic plug & work robot operation. The XBot2 middleware, described in Section 5.1, allows for a seamless transition (from the user perspective) between different robot configurations and provides a uniform interface to control the robot for high-level tasks. The software components for implementing such high-level tasks are presented in the following sections: for BIM-based mission specification in Section 5.2, for autonomous navigation in Section 5.3, for drilling in Section 5.4 and for collaborative transportation in Section 5.5.

## 5.1 | The XBot2 Middleware

The core part of CONCERT's software stack is formed by the *real-time middleware*, that is the lowest software layer which allows to run user code, under real-time constraints. The XBot2 real-time middleware for reconfigurable robots Laurenzi et al. (2023) was designed specifically

---

[||] https://github.com/ADVRHumanoids/modular_hhcm

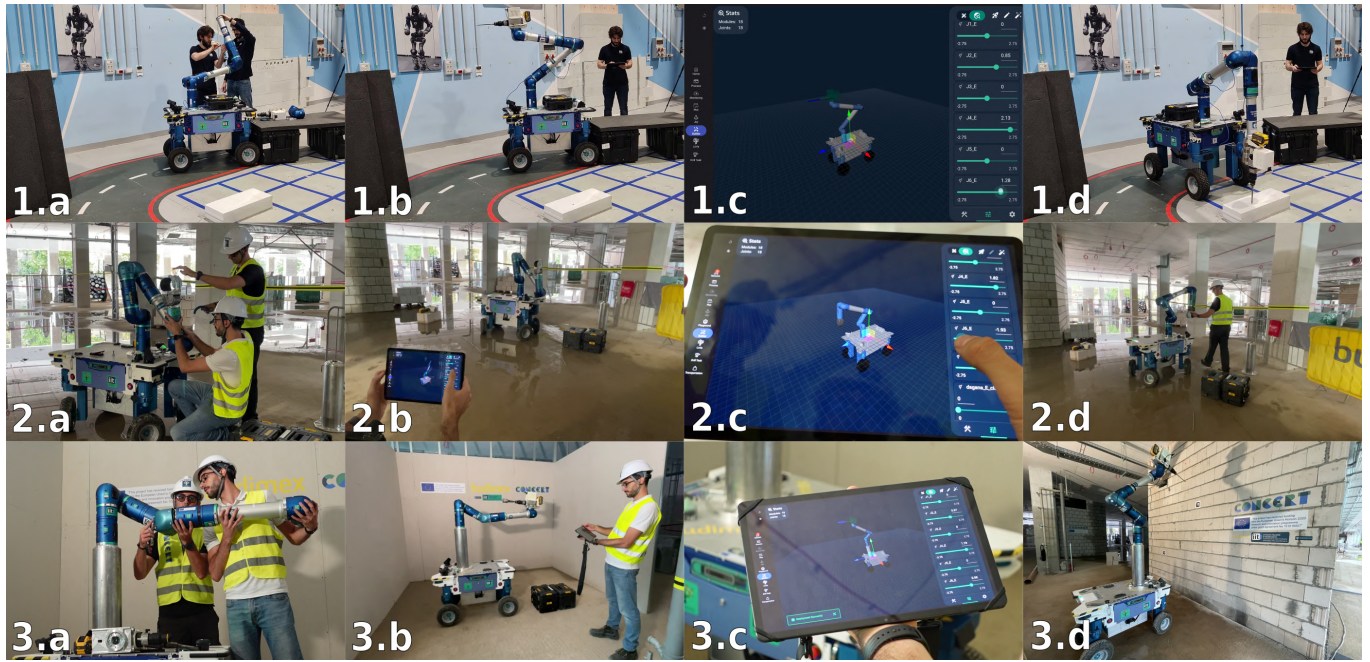

**FIGURE 7** Three examples of model reconstruction experiments are shown for three different morphologies in different scenarios. For each of them: the modular robot is manually assembled by the users (phase a); once assembly is finished the topology recognition procedure is started from the user interface on the tablet (phase b); the topology is recognized and the model reconstructed as can be seen from the GUI, the user can manually customize some options like homing configuration or end-effector addons (phase c); finally, with the reconstructed model available, the robot is ready to perform high-level tasks such as drilling, collaborative transportation, etc. (phase d)

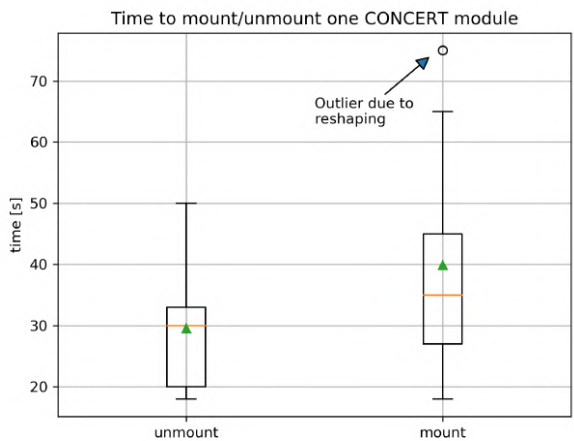

**FIGURE 8** Statistical distribution for the mounting and unmounting time of a CONCERT module, evaluated on all modules that comprise our library. The orange line and the green triangle mark the median and the mean, respectively. Outliers in the mounting distribution are due to the additional time required to move one or more joints to ease the mounting of the following modules.

for CONCERT based on the following requirements. It must provide (i) convenient hardware abstraction capabilities that are suited for the reconfigurable robot use case, including automatic detection of the available hardware, and APIs to discover all the available devices; (ii) a flexible multi-threaded architecture to allow different modules (a.k.a. *plugins*) to run at different frequencies, without placing any burden on the user to implement explicit synchronization mechanisms (e.g. mutual exclusion, condition variables, etc.); (iii) operating system abstractions that allow the framework to run on different platforms (Vanilla Linux, Linux/PREEMPT_RT, Xenomai, and others); (iv) a minimal inter-plugin communication and configuration tool-chain to promote *modularity* and *re-usability* of plugins, such as real-time safe publish/subscribe, remote function call (RFC), and global parameters.

The hardware abstraction layer of XBot2 is configured via YAML-based configuration files in addition to standard descriptions of the robot kinematics such as URDF, SRDF. All these files can be automatically generated from information retrieved during the network scan phase.

## 5.2 | BIM-integrated mission parametrization and execution

The concert software architecture contains components for the generation and parametrization of robot missions from a user interface based on Building Information Modeling (BIM) data. The user interface of CONCERT is based on our prior work in Terzer et al. (2024), where we presented a facilitated programming approach that allows non-expert users to graphically interact with a BIM model, regulate mission parameters, and finally dispatch a mission as a behavior tree (BT).

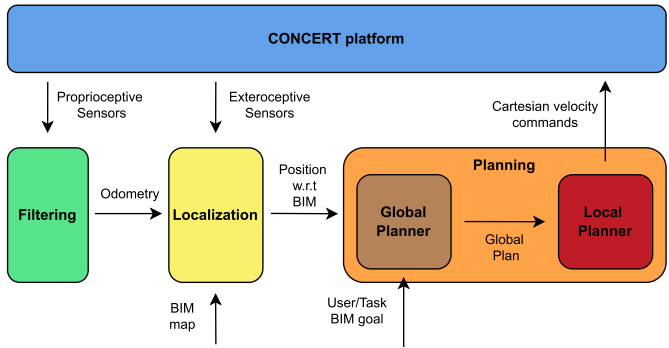

**FIGURE 9** Graphical representation of the interaction between autonomous navigation components

In the CONCERT software architecture, we use ROSBIM[**] our developed library, whose early version has been introduced in Giusti et al. (2021), to parse BIM models in the form of IFC files, which in turn must be provided a priori. The current version of ROSBIM comes with a plugin-based architecture that allows the use of plugins to interface with the BIM model. For the execution of the drilling mission we use: an ExportMap plugin to export an occupancy grid map from a slice of a building floor and use it for robot navigation, and an ExportGeometries plugin to export objects of specific types (e.g. Walls, Doors, Windows) and use them in the user-interface for environment visualization. When the users interact with the BIM model, they can select a goal on a wall. In response, the BIM model provides the mobile base pose in front of the selected wall as a tuple $< x, y, \theta >$, where $x$ and $y$ are the Cartesian coordinates and $\theta$ represents the rotation around the $z$-axis relative to the navigation map frame. This mobile base pose is then used by the autonomous navigation module to position the robot in front of the wall. Meanwhile, the drilling area frame serves as the Cartesian reference for the whole-body planner, which positions the end-effector to have a clear view of the drilling points.

## 5.3 | Autonomous Navigation

As illustrated in Figure 9, our developed autonomous navigation toolchain is composed of the following components.

- *filtering*: allows us, on the robot, to estimate the Cartesian velocity of the platform computed through the wheel state information given by the proprioceptive perception modules Moore and Stouch (2016). The output is the odometry, position on the plane and orientation of the mobile-base.

- *localization*: estimates the platform position inside the BIM map. Here we used LIDARs as exteroceptive sensors and the odometry data.
- *planning*: is composed of the global planner to compute the desired global paths to reach the input desired goal, and the local planner to control the platform make it follow the global plan. Here we adopted the *navigation2* Macenski et al. (2020) tool for planning.

Due to the strong relation between the navigation components presented and the platform mobile-base module arrangement, the CONCERT platform navigation must be adapted with respect to its user-defined configuration. We perform this adaption automatically as in Morganti et al. (2024) through a procedure for automatic vehicle kinematics recognition and navigation deployment after assembly. This ensures the reconfigurability of the modular CONCERT platform, a navigation software structure that is flexible and capable of adapting to the different configurations of the mobile-base modules. In particular, it relies on the automatic deployment of navigation toolchain starting from the analysis of the generated URDF of the platform assembly. Here, we recognize the possible vehicle kinematics by considering the wheels mounted on and we configure the perception capabilities by looking at the available perception modules. Such approach opens the possibility for the user to reconfigure the mobile base assembly without needing the support of an expert for deployment of the navigation.

## 5.4 | Planning and Control for Drilling

In the following, we describe the planning and control framework for the execution of the drilling tasks. For this task, the CONCERT robot is asked to drill a sequence of holes in correspondence of markers applied by the user in the form of circular blue blobs drawn using a permanent marker and ArUco markers. With these two different marks, we aim to replicate two prevalent scenarios in a construction setting. One involves the operator manually indicating the drilling location by hand using a marker, while the other entails simplifying the process by encoding a pattern in an Aruco marker, such as a sequence of equidistant holes, eliminating the need for time-consuming measurement and marking of each individual hole.

The generation of trajectories for reaching the drilling area and performing the task involves additional considerations beyond a straightforward Inverse Kinematics (IK) approach due to the high contact force and resulting torque demands involved. Construction sites are complex environments whose characteristics change rapidly with the progress of the work and the robot has to adapt to these extremely dynamic environments. Further, the modular nature of CONCERT requires extra care of torque limits and self-collision since the robotic arm can reach high extensions majorly stressing the motors at the beginning of the kinematic chain. Moreover, accessing drilling locations poses kinematics challenges (see Fig. 10), and intricate paths are frequently necessary to navigate toward the target, mitigating the risk of collisions and avoiding singularities that could impede the drill bit's successful penetration.

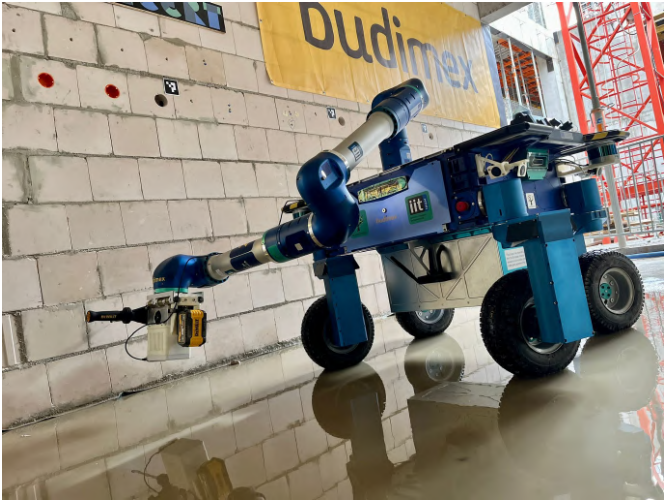

**FIGURE 10** CONCERT performing a drill $0.3\,\text{m}$ from the ground.

For the aforementioned reasons, the planning framework has been designed based on a joint-space sample-based planner that generates whole-body trajectories to approach the drilling spot depending on the environment state perceived as an OcTree through the embedded Lidars and Intel Realsense depth cameras. To avoid unfeasibilities, the planner is constrained by geometric (i.e., (self-) collision avoidance) and kinodynamic constraints (i.e., joint and torque limits) and all the sampled states undergo these constraints to assess their validity. The finishing drilling motion is instead accomplished by end-effector trajectory obtained through the IK controlled through a Cartesian impedance controller to mitigate the effect of impacts and interactions with the environment during the execution of the task. The problem of planning and executing a drilling task is performed following the steps below:

1. Acquisition of the surrounding environment using the embedded perception system;
2. Selection of the drilling areas directly clicking on the sensed point cloud or from a BIM model, if available;
3. Planning of whole-body trajectories to approach each clicked area;
4. Execute the planned trajectory;
5. Execute the drilling tasks through a closed-loop controller.
6. Repeat steps 4) and 5) until all tasks are performed.

We define a *drilling area* as a portion of the environment where one or more blobs/ArUco markers are observable when the robot's camera is at a specific distance $d$, and *drilling point* as the point in a drilling area where a blob is detected or as one of the points in the set encoded by the ArUco marker. Each of the $N_{DA}$ drilling areas is associated with a frame $\mathcal{F}_{DA}^{i}$, provided by the user, and each of the $N_{DP}^{(i)}$ drilling point in the $i$-th drilling area is identified by a frame $\mathcal{F}_{DP_j}^{(i)}$. Both the drilling area and the contained drilling point frames have the z-axis aligned with the surface normal $\boldsymbol{n}_i$.

The framework to perform the drilling tasks is implemented as a behavior tree graphically represented in Figure 11. In this scenario, the drilling areas frames $\mathcal{F}_{DA,i}$ are provided to the robot through a BIM model or by directly clicking them within the sensed point cloud in case the BIM model is not available. The BIM model also provides a base goal pose expressed as a tuple $< x, y, \theta >_i$ that moves the robot's base in front of the relative drilling area through the autonomous navigation module (Sec. 5.3). When the BIM model is not available, the user selects the drilling areas directly clicking on the point cloud sensed by the embedded camera system. This implementation is schematized by the *initialization* module in the behavior tree of Fig. 11. In this second case, due to the limited maximum distance of the camera, which ranges around $2\,\text{m}$ and $3\,\text{m}$, the robot is assumed to be already in front of the wall where the drilling task has to be performed, thus skipping the autonomous navigation. Then, the algorithm maps each $\mathcal{F}_{DA,i}$ into a sequence of whole-body robot configurations through the Null-Space Posture Generator (NSPG) while contemporary planning and executing a sequence of whole-body trajectories connecting the configurations just generated. Lastly, upon completion of each trajectory's execution, with the robot in front of the $i$-th drilling area, it drills all the $N_{DP}^{(i)}$ drilling points it is able to detect in that configuration. The robot's configuration generation, plan, and execution of the whole-body trajectories is carried out in parallel to minimize the dead time and the completion time. The rest of this section will go through every components of the planning framework, highlighting the role of each block in the autonomous generation of whole-body motions to fulfill the task.

### 5.4.1 | Environment and Drilling Area Perception

In case a BIM model is not available, the framework is initialized with the acquisition of the environment through the embedded perception system in the form of a point cloud and RGBD images. Among all the sensors available on CONCERT, for the drilling task we used the two Lidars mounted on the base, which returns a non-colored point cloud, and the depth camera mounted on the drill, returning a colored point cloud and a RGBD image instead. Since multi-camera information leads to overlapping and redundant measures, the acquired point cloud is processed and transformed in a OcTree that removes similar points using a statistical approach. To reduce the sensor noise, which is particularly evident on the Intel Realsense camera at distances higher than approximately two meters, progressive point clouds acquisitions are averaged in time relying on the position in pixel of each point. The resulting map is then provided to the planning layer as an occupied region in space so that the robot has to avoid during the generation of the approaching trajectory. At this point, the user is provided with a full description of the environment and can select the areas to be drilled by clicking directly on the point cloud.

An intermediate process is in charge of generating a frame $\mathcal{F}_{DA,i}$ aligned with the surface normal $\boldsymbol{n}_i$ passing through the $i$-th drilling area picked by the user. This is obtained by first performing a point cloud downsample to speed up the computational efficiency of the algorithm, and then by averaging the normal vectors related to all the points around the clicked ones in an empirically chosen radius of $0.07$ m leveraging on the Point Cloud Library Rusu and Cousins (2011). Then, the frame is constructed aligning the local $z$-axis with the estimated normal and it is provided as a goal reference to the planner.

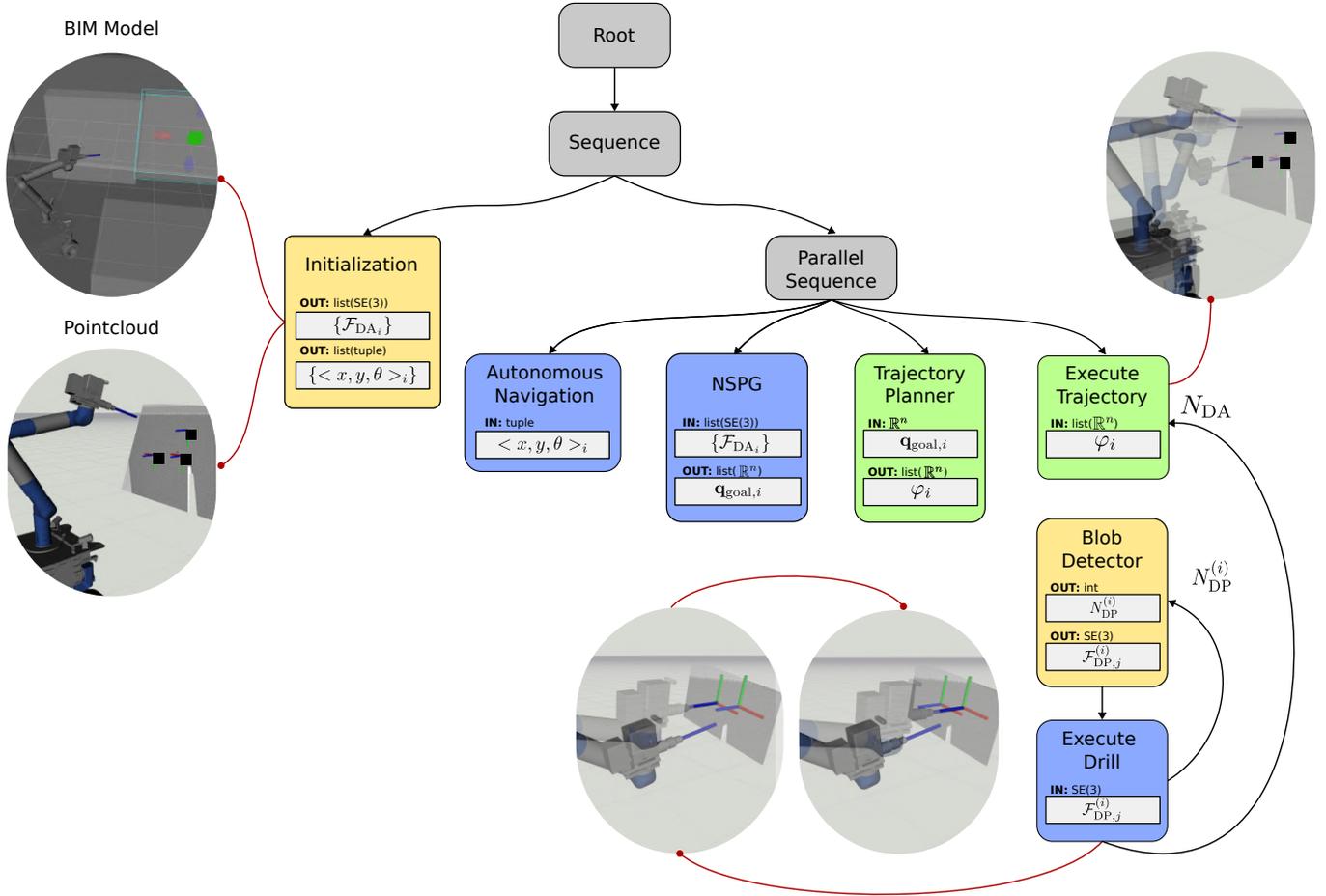

**FIGURE 11** A graphical representation of the behavior tree used to execute autonomously the drilling tasks. Yellow actions correspond to those actions in which perception is involved, while blue and green actions are the ones performed in Cartesian and joint space respectively. Black arrows represents the flow of the actions, and red lines connect each block with a figure describing the scene.

### 5.4.2 | Sample-Based Planner

Given the current robot configuration, a map of the surrounding environment, and a set of $N_{DA}$ goal frames $\mathcal{F}_{DA,i}$, the framework triggers the planner generating whole-body trajectories to drive the robot in front of the drilling areas. We consider the CONCERT robot as a floating-base robot whose configuration is fully described by a vector $\boldsymbol{q} \in SE(3) \cdot \mathbb{R}^n$ with $n$ being the number of DoFs. Thus, the *configuration space* $\mathcal{Q}$ is the space of all robot configurations into which the planner will sample states to generate a feasible robot motion, and it will have dimension $\dim(\mathcal{Q}) = \dim(\boldsymbol{q})$. Assuming that the robot moves on flat terrains only, the motion on the $z$-axis and the roll and pitch rotations of the base is constrained and the base motion can be reduced and simplified to a planar motion in an $SE(2)$ space. Further, the configuration space can be split in two sub-spaces $\mathcal{Q}_{free} \subseteq \mathcal{Q}$ and $\mathcal{Q}_{obs} \subseteq \mathcal{Q}$ that contains all the configurations in free and occupied space respectively so that $\mathcal{Q}_{free} \cap \mathcal{Q}_{obs} = \emptyset$. Lastly, $\mathcal{Q}_{free}$ can be split in $\mathcal{Q}_{feas} \subseteq \mathcal{Q}_{free}$ that contains all the configurations that respects all the other geometric and kinodynamic constraints such as self-collision avoidance, Cartesian constraints, and joint limits, and $\mathcal{Q}_{unfeas} \subseteq \mathcal{Q}_{free}$ of all the unfeasible configurations.

With these quantities defined, given a start and a goal configuration $\boldsymbol{q}_{start}, \boldsymbol{q}_{goal} \in \mathcal{Q}_{feas}$, the motion planning problem can be formalized as the generation of a continuous path $\varphi : [0,1] \to \mathcal{Q}_{feas}$ that connects $\boldsymbol{q}_{start} = \varphi(0)$ to $\boldsymbol{q}_{goal} = \varphi(1)$ (Kingston et al. 2018). To solve the motion planning problem, the planner grows a search tree $\mathcal{T}$ seeded in $\boldsymbol{q}_{start}$ and uses a sample-based algorithm to samples robot's states from $\mathcal{Q}_{feas}$ with a specific logic that will make the tree grow towards the goal region (LaValle 2006). The tree expansion and the configuration space exploration continues until the $\boldsymbol{q}_{goal}$ can be directly connected with $\mathcal{T}$, meaning that a feasible robot motion has been successfully generated. In our planner, we use a variation of the RRT algorithm called RRT-Connect that incrementally builds two RRTs rooted at the start and goal configurations (Kuffner and LaValle 2000a), thus enhancing the performances of the planner in high dimensional configuration spaces.

Once the user provides the set of goal frames $\mathcal{F}_{DA,j}$ by clicking on the perceived point cloud, the next step consists on mapping the Cartesian goals in whole-body goal configurations $\boldsymbol{q}_{goal,j}$ to be provided to the planner. Specifically, the robot has to align the drill axis with the surface normal in a well-defined point in space while respecting the kinematic limits and the collision safeness. Additionally, the algorithm must generate configurations far from any singularity and as close as possible to $\boldsymbol{q}_{start}$ to reduce the exploration required by the planner and the length of the trajectory connecting the two states.

The generation of a feasible and connectable configuration is a crucial point for a sample-based planning framework that has been already addressed in Rossini et al. (2021) presenting the Null-Space Posture Generator (NSPG). The NSPG seeks for a feasible configuration by altering a nominal configuration in the null-space of some Cartesian and geometric constraints. In the case of interest, we ask the NSPG to generate the goal configuration starting from $\boldsymbol{q}_{start}$ and moving in the null-space of the Cartesian task compliant with $\mathcal{F}_{DA,j}$, rejecting any robot pose that is in collision or cannot generate a trajectory that moves the trajectory towards the wall. Specifically we control the Cartesian position of the camera frame mounted on the drill $\mathcal{F}_{cam}$ at a specific distance $d = 0.6$ m from the wall, and oriented with the camera axis parallel to the surface normal.

Every time a new goal configuration $\boldsymbol{q}_{goal,j}$ is available, the planning routine starts in parallel with the generation of the remaining ones in a First In First Out (FIFO) order. The procedure starts by taking the first available $\boldsymbol{q}_{goal,j}$ and generating a whole-body trajectory $\varphi_j$ root in $\boldsymbol{q}_{start,j}$. The starting configuration is taken reading the joint position from the robot. Following the same FIFO order, the planned trajectories are executed in parallel with the planning procedure. By executing each trajectory $\varphi_j$, the robot drives to the $i$-th drilling area with a clear view of all the $N_{DP,j}$ drilling points. Further, with a closer view and a more reliable point cloud of the $i$-th drilling area, the estimation of the surface normal $\boldsymbol{n}_i$, as well as the drilling points poses $\mathcal{F}_{DP,j}^{(i)}$, can be carried out in a more precise way.

### 5.4.3 | Closed-Loop Drilling

When the $i$-th drilling area is reached, the robot detects and drill all the $N_{DP}^{(i)}$ drilling points visible. Specifically, the blue markers are now detected using RGBD data from the camera exploiting the OpenCV library (Bradski 2000) projecting the segmented blue pixel that forms a circular shape onto the depth cloud. Alternatively, in case the robot detects an Aruco marker, the robot automatically sets the drilling point frames $\mathcal{F}_{DA,j}^{(i)}$ following a pattern associated with the Aruco's id. Once in the correct position and orientation, the drilling task is performed commanding a Cartesian velocity $v_{drill}$ along the axis of the drill bit. To reject external disturbances along the perpendicular axis and guarantee a compliant behavior of the drill during the penetration, the robot arm has been controlled in a Cartesian impedance mode. Simultaneously, the position of the drill bit is continuously adjusted in a closed-loop fashion using visual servoing. This process aims to minimize any misalignment in the plane perpendicular to the drill axis and to correct the alignment of the drill bit with respect to the surface normal. During each iteration, $\mathcal{F}_{DP,j}^{(i)}$ pose is adjusted based on the estimated pose of the blue blob, orienting the z-axis with the surface normal $\boldsymbol{n}_i$. The estimation involves segmenting the blue pixels from the captured image and determining the blob's location and dimensions (in pixels) using OpenCV's[††] blob detection logic. By combining this information with data from the depth camera, the 3D position of the blob and the orientation of the surface normal can be derived. Assuming that the z-axis of the drill bit frame $\mathcal{F}_{db}$ is aligned with its rotation axis, the closed-loop controller aligns the drill with the blob. This is achieved by computing the homogeneous transform from $\mathcal{F}_{db}$ to $\mathcal{F}_{blob_N}$, eliminating its x-y components, and aligning the z-axis with the estimated surface normal in the opposite direction. In essence, the goal is to calculate a Cartesian reference for the drill bit in the form:

$$^{db}\tilde{\boldsymbol{T}}_{blob_N} = \begin{bmatrix} & & & 0 \\ \boldsymbol{x} & \boldsymbol{y} & -\boldsymbol{n}_i & 0 \\ & & & \hat{z} \\ 0 & 0 & 0 & 1 \end{bmatrix} \tag{3}$$

Here, $\hat{z} = z_0 + v_{drill} \cdot dt$ denotes the incremental reference on the drill axis. This reference is computed from the current drill bit position $z_0$ and the Cartesian velocity $v_{drill}$ along the drill axis integrated over the controller time period $dt$. The direction cosines $\boldsymbol{x}$ and $\boldsymbol{y}$ can be arbitrarily assigned since the closed-loop controller primarily influences the z-axis components and x-y position.

### 5.5 | Collaborative Transportation

A critical objective of the CONCERT project is to provide a platform that can actively collaborate with human operators to transport objects whose dimensions and weight would instead require two or more workers. For this purpose, we developed an interface, starting from the preliminary results presented in Muratore et al. (2023), to streamline the collaborative transportation tasks: a set of transportation modes allows the user to select the desired action via a portable GUI or by sending vocal commands through a headset connected to the robot's PCs. This set of control modes includes:

- a *follow-me* mode, where the robot wheels are driven by the force applied to the end-effector by human interaction,
- a *teaching* mode, where the robot arm can be hand-guided by the user to the desired posture,
- an *autonomous* mode, where the robot accepts remote commands without any interaction with the user,
- a *stop* mode, so that the operation can be safely stopped.

---

[††] https://opencv.org/

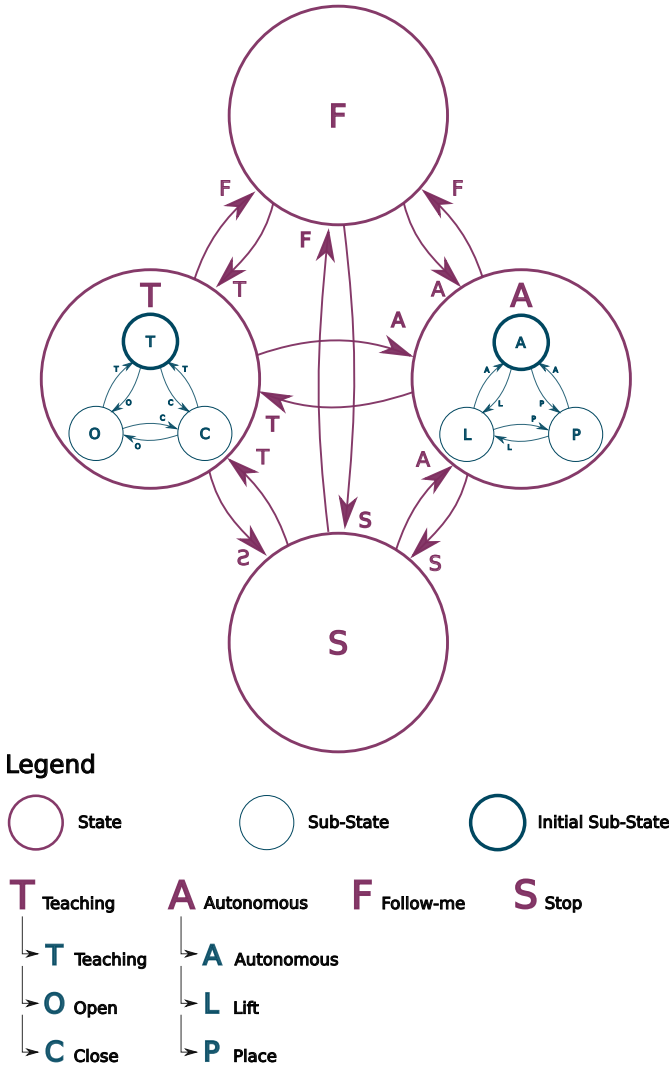

**FIGURE 12**  The collaborative transportation task state machine.

Figure 12 shows a finite state machine diagram where the 4 main modes are represented as states. States and actions to switch between states share the same name for the sake of simplicity.

Follow-me mode refers to the collaboration task in which the human operator is leading the robot by hand via the force exerted on the CONCERT arm: the estimated force at the end effector is translated into a base velocity reference that is sent to an omni-steering module that maps those velocities into wheel steering and rolling velocities. Oppositely, in the teaching mode, the mobile base is kept fixed, and the arm is controlled by compensating the gravity, allowing the operator to safely move the end-effector to the grasping configuration.

In teaching mode, the robot can perform the *open* and *close* actions to control the gripper, while in autonomous mode, it can execute *lift* and *place* actions to handle the payload. These actions correspond to substates within the teaching and autonomous states, respectively. When entering the teaching state, the robot initially defaults to a general teaching sub-state but transitions to the open or close sub-state depending on the requested action. Similarly, in the autonomous state, the robot switches between the lift and place sub-states as needed, as illustrated in Figure 12. All state transitions are triggered by vocal commands with names matching the desired actions, which the robot recognizes in collaborative transportation mode. To issue a vocal command, the user must prepend it with the prompt "Hey CONCERT." For example: "Hey CONCERT, Teaching".

Figure 13 illustrates the various states involved in a collaborative transportation task experiment. For this use case, we set up a robotic arm with seven DoFs and one $40\,$cm Passive Link module to grasp objects from the ground. The kinematic chain starts with the Straight A - Elbow A -Straight A Joint modules. Then, the Elbow A - Straight B Joint modules are mounted followed by the Passive link module and a two-DoFs wrist made by the Elbow B - Straight B Joint modules. The seventh DoF enhances redundancy, enabling the robot to track the user's movements more smoothly. At the end of the kinematic chain, one of the presented Active Grippers is mounted depending on the grasped object. In this configuration, the total length of the robotic arm is $2.0\,$m mounting an End-Effector module that weights $2.5\,$kg, which will manipulate objects up to $20\,$kg together with the human operator.

The process begins with the robot in the *teaching* state, where the user can specify the position of the object to transport. Using corresponding commands, the user can close the gripper and lift the object from the ground. Since *lift* is a sub-state of the *autonomous* state, the main state must change before lifting. To reduce verbosity, saying "Hey CONCERT, Lift" while in the teaching state automatically switches the robot to the *autonomous* state and initiates the *lift* action. Once the object is lifted, the user can switch to *follow-me* mode to guide the robot to the desired location. Upon arrival, the command "Hey CONCERT, Place" will both switch to the *autonomous* state and trigger the *place* action. The user can then return to the *teaching* state to open the gripper. Note that certain actions are restricted in specific states to prevent errors. For example, the open command is ignored in *autonomous* or *follow-me* mode to avoid accidental drops. In this experiment, after placing the object, the user reactivates *follow-me* mode to park the robot in another location. The *teaching* and *close* commands are then used to instruct the robot to hold a structure needed for assembly. By setting the robot to the *autonomous* state, it remains stationary while the user completes the task.

## 5.5.1 │ The Follow-me MPC Formulation

To execute this task, we decided to use an anthropomorphic arm with 7 DoFs to make the maneuverability of the arm as similar as possible to human capabilities. The control problem related to the follow-me mode is formulated as a Model Predictive Control (MPC), which takes the force readings from the end-effector as input. The output of the MPC is the base velocity reference that is sent to the wheel controller. The MPC includes:

- joint velocity and acceleration regularization;
- robot postural for a desired joint configuration;

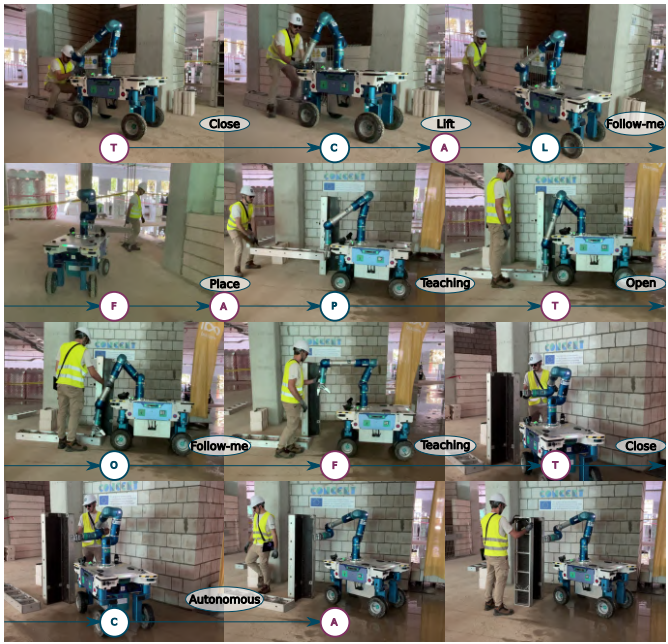

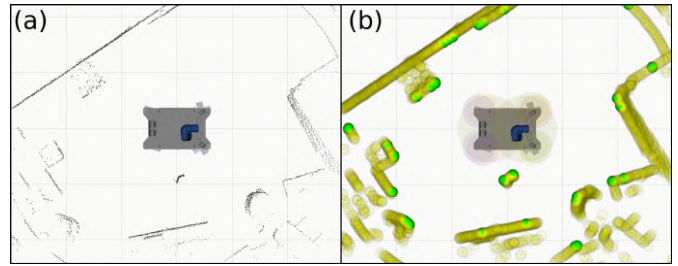

**FIGURE 14** (a): the grid map produced by the sensors. (b): the grid map is processed to generate the obstacles. The yellow and green spheres are "inactive" and "active" obstacles, respectively. The spheres on the robot are the repulsive areas for the lidars and the sonars.

## 5.6 | Obstacle avoidance

An obstacle avoidance module was developed for the CONCERT robot, which enhances the navigation skills provided by the omni-steering controller in the hand-guiding mode, by incorporating detection and autonomous avoidance of obstacles in the proximity of the robot: the base velocity reference sent to the omni-steering module is modified to steer the robot along obstacle-free directions.

The problem described in Subsection 5.5 for the hand guiding mode is extended to take as input the obstacles' position from the environment, continuously scanned by the lidar and sonar sensors. This is achieved by augmenting the MPC in 5.5 with a suitable term, i.e. a soft constraint that imposes a minimum distance between the obstacles the robot:

$$\|\boldsymbol{o}_{robot}^k - \boldsymbol{o}_{obs}^k\|^2 \le (r_{obs} + r_{robot}) \tag{5}$$

where $\boldsymbol{o}_{robot}^k$ and $\boldsymbol{o}_{obs}^k$ are the origin of the collision spheres of the robot and the obstacles, while $r_{robot}$ and $r_{obs}$ are their respective radius. The output of the MPC is once again the base velocity reference but modified such that CONCERT follows the human guide while autonomously adjusting its motion to avoid obstacles.

The environment is mapped using two LIDARs, which generate a point cloud, and eight sonars, which detect obstacles in the robot's vicinity. All the sensor's data is transformed into a grid map, as depicted in Figure 14(a), where each grid cell is classified as either empty or occupied. To model occupied cells, spheres of a defined radius are used, as shown in Figure 14(b). Similarly, the robot is represented using a collection of spheres: two spheres approximate its rectangular perimeter for the lidar map, while individual spheres, centered on each sonar, represent the coverage areas of the sonar sensors.

As obstacle spheres approach the robot spheres, the MPC problem dynamically adjusts the robot's trajectory to ensure the spheres remain at a safe distance, effectively guiding the robot away from obstacles that breach a predefined proximity threshold.

When no obstacles are nearby, the robot's base velocity aligns with the direction of the force applied to the end-effector, enabling seamless following of the human operator. However, if an obstacle encroaches into the robot's repulsion zone, the MPC controller modifies the base

**FIGURE 13** The different stages in the collaborative transportation task. For each stage, the corresponding state of the state machine is indicated, and for every transition, the command that triggers it is specified. Additionally, for commands that result in multiple state transitions, the intermediate states are also depicted.

- minimization for the desired Cartesian reference for the end-effector and base;
- state and input bounds;
- non-flight constraints.

The MPC minimizes the costs while constraining the solution: the state and input bound ensure that the joint position, velocity, and acceleration reside inside the feasible region, while the non-flight constraints guarantee that the robot can only move in the horizontal plane.

A mass-damper system is used to impose the desired dynamics of the end-effector, establishing a relationship between the input force and end-effector position and velocity. The dynamics are described by the equation:

$$\boldsymbol{M}\ddot{\boldsymbol{x}} + \boldsymbol{D}\dot{\boldsymbol{x}} = \boldsymbol{F}_{ee} \tag{4}$$

where $\boldsymbol{M}$ and $\boldsymbol{D}$ are the desired virtual mass and damping matrices, $\ddot{\boldsymbol{x}}$, $\dot{\boldsymbol{x}}$ are the acceleration and velocity of the virtual system and $\boldsymbol{F}_{ee}$ is the force exerted at the end effector. Then, the desired velocity sent to the robot base is yielded by integrating the system state using the end-effector force as an input. The integrated state over the optimization horizon of the MPC is sent as a reference to the Cartesian end-effector position, and the solver will minimize the distance between the current end-effector position and the desired one.

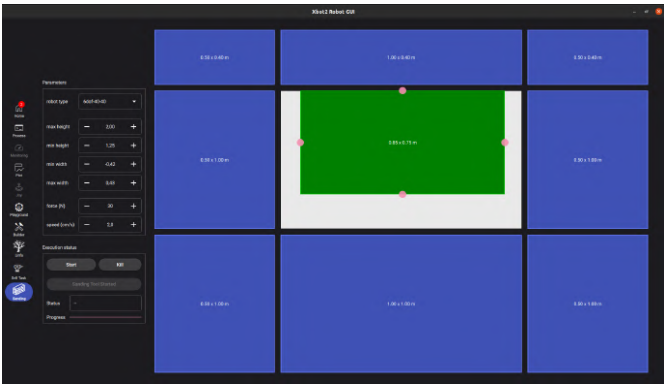

**FIGURE 15**  The CONCERT sanding and spraying GUI.

velocity to prevent collisions. It is important to note that both lidar and sonar sensors are essential for effective obstacle avoidance, as the lidar sensors mounted on the robot have a minimum detection range beyond which obstacles are no longer detected.

To optimize computational efficiency, the grid map is processed to focus only on the obstacles closest to the robot, which are then incorporated into the MPC problem. As illustrated in Figure 14, only "relevant" obstacles (displayed in green) are activated for the calculations, while others (shown in yellow) are disregarded. This is achieved by dividing the map into sectors—radial slices centered on the robot's base—where only the nearest obstacle within each sector is considered active.

## 5.7 | Sanding and Spraying

The last two use cases studied and developed in the CONCERT project are related to the sanding and insulation spraying of a vertical wall. These two tasks share some similarities that allowed us to tackle them using a similar approach. Indeed, both tasks require the robot to cover an arbitrarily wide area of a vertical wall moving the end-effector smoothly and uniformly along the surface to guarantee a uniform result. First, we designed and built two modular end-effectors for the execution of the two tasks: The Sanding Tool and Spraying Tool modules, which have been described in Section 3

Differently from the drilling use-case for sanding and spraying we developed a solution that doesn't rely on the BIM information to perform the task. This is to allow the execution of these operations also when the BIM is not available or incomplete. Also, operations like sanding are performed as a last step and the current constructed building might have diverted from the original plan.

The selection of the sanding/spraying area is performed, therefore, in a local way (with respect to the robot's current position) and is specified by the user from the GUI. In Figure 15 is shown the layout of the GUI, from where the user can select the area of the wall where to perform the task. Since the sanding and spraying tasks cannot be interrupted is important to ensure the required trajectory can be tracked continuously by the robot without any self-collision. To achieve this, the wall has been divided into predetermined areas the user can select that will result in the robot moving to a specific distance from the wall and assuming a specific posture that ensures the area to be covered is completely inside the robot workspace without any self-collision or discontinuity. The lateral panels of the GUI are thought to be used when you are in proximity to a lateral wall, and therefore, we have an additional constraint of avoiding collision with it. This division in predetermined areas allows us to avoid moving the robot base during the task execution and instead move it only when changing areas. In fact, since for sanding and spraying tasks, precision is paramount, we preferred not to rely on the tracking of the wheels, which are designed for outdoor work, and rely only on the arm, which guarantees more accuracy.

Parameters such as the maximum and minimum height at which the task can be performed will be inferred from the automatically discovered morphology of the robot currently used. Other parameters are instead selectable from the GUI by the user, such as the reference force to be applied by the robot during the sanding task or the desired end-effector speed. Also, each of the sanding/spraying predetermined areas can be easily reshaped from the GUI in case a smaller area should be covered due to the presence of any "obstacle" on the wall (e.g. columns, windows, doors, etc.).

When the user starts the mission after selecting the desired area the robot first aligns to the wall and approaches it at the specified distance. While in the spraying case this distance is maintained for the whole task, in the sanding case, before beginning the motion, the arm also establishes contact with the wall and set the correct force to be applied. Thanks to the torque sensors present on CONCERT joints the external forces at the end-effector can be estimated and regulated. Once the required force setpoint is reached, the robot waits for the user input to start the sanding motion, since the tool can be activated only manually in the current implementation and its activation should be synchronized with the start of the trajectory. In both the sanding and spraying cases, the robot will follow a path that ensures the whole desired area is covered: a serpentine trajectory is generated where the number of segments depends on the width/height of the rectangle specified in the GUI and the width of the spray cone (or diameter of the sanding tool in the sanding case).

In both cases, a cartesian impedance controller as the one implemented in Romiti et al. (2022) has been used. In the sanding case, a simple PD regulation controller has been applied on top of the impedance controller to modify the desired position of the end-effector along the direction perpendicular to the tool sanding surface ($z_{ee}$). Modifying the controller in this way implements an *admittance* control law (Ott et al. (2010)) for the direction $z_{ee}$ while in the other directions, the usual impedance behavior is maintained.

## 6 | EVALUATION

This section contains the experimental results and a performance evaluation of CONCERT using the software tools described in the previous

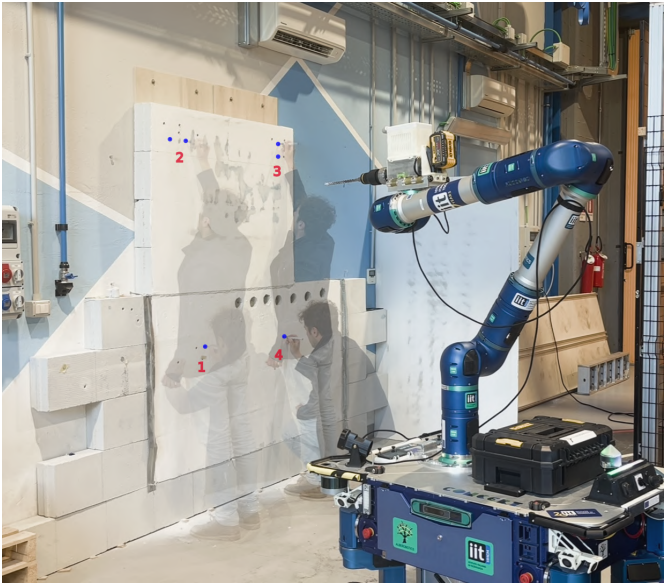

**FIGURE 16** Screenshots from the initial stage of the experiments where the user draws the blue blobs that will be used as a reference for the drilling task by the robot. The user marks the wall in four drilling areas numbered in red. Each blue marker is highlighted with a blue circle.

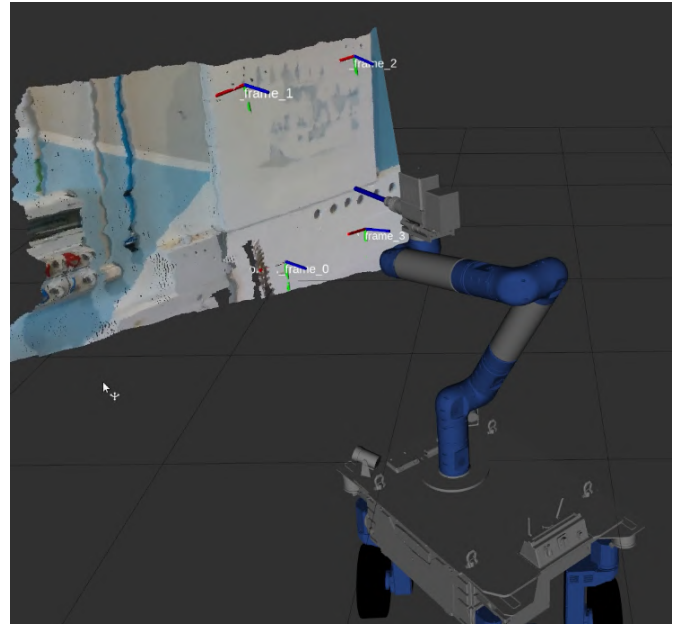

**FIGURE 17** Robot's view of the wall after the user has clicked on the perceived point cloud to mark the centers of the drilling areas $\mathcal{F}_{DA}^{(i)}$

section. All the following results, have been collected in a real construction site.

## 6.1 | Drilling

\* The proposed planning framework was validated through simulations and tested on a real robot in complex environments across three scenarios: drilling all visible points within a designated drilling area selected from (i) a point cloud, (ii) a BIM, and (iii) a hole pattern starting from an ArUco marker. For these tasks, the robot is assembled to work with a six DoFs configuration ending with the Drill End-Effector. Depending on the number and arrangement of Passive Link modules, three distinct morphologies can be obtained, labeled A, B, and C. The kinematics of Morphology A begin with the *Straight A - Elbow A - Straight A* Joint modules, followed by the *Elbow A - Straight B* Joint modules. To extend its workspace, a passive $0.4\,\mathrm{m}$ Link module is added before the final *Elbow B* Joint module, which houses the Drill End-Effector module. In this configuration, the arm reaches a total length of $1.75\,\mathrm{m}$, supporting an end-effector weighing approximately $10\,\mathrm{kg}$. Morphology B extends the arm to $2.15\,\mathrm{m}$ by integrating a second $0.4\,\mathrm{m}$ Passive Link module after the third Joint module. Morphology C further expands the reachable workspace by incorporating the Passive Base Link module before the first Joint module, increasing the arm's total length to $2.75\,\mathrm{m}$.

For the first two cases, each drilling area is associated with a frame, denoted as $\mathcal{F}_{DA}^{(i)}$. The robot plans a whole-body trajectory to move the drill parallel to the surface normal, maintaining a specified distance from the wall, allowing for precise detection of drilling points $\mathcal{F}_{DP,j}^{(i)}$. The size of the drilling area depends on the camera's field of view and its distance from the wall. Assuming a field of view of $2\theta = 55°$, which is the minimum between the horizontal and vertical fields of the Intel® RealSense™ D435i camera, and a distance of $d = 0.6\,\mathrm{m}$ from the wall, the radius of the drilling area can be calculated as:

$$\rho = d \cdot \tan\theta \approx 0.33\,\mathrm{m} \qquad (6)$$

This means that a set of drilling points belongs to the same drilling area if their distance from the area center $\mathcal{F}_{DA}^{(i)}$ is lower than the radius $\rho$.

In the first scenario, the user draws six blue dots in four clusters, each forming a drilling area as depicted in Figure 16. The lowest blobs are drawn at $0.95\,\mathrm{m}$ from the ground, while the highest ones are centered at $2.15\,\mathrm{m}$ from the ground with a relative distance of $\approx 0.1\,\mathrm{m}$ between one another. Since the robot must reach different regions of its vertical workspace, we use morphology B for this use case. Once drawn the drilling points, the user can indicate the robot the centers of the drilling areas by clicking on the colored point cloud perceived by the robot in correspondence of the visible blue dots (see Fig. 17). In this scenario, we assume that the BIM is not available and the behavior tree is initialized computing the drilling area frames $\mathcal{D}_{DA,j}$ from the point cloud perceived by the drill camera, skipping the autonomous navigation stack. Figure 18 collects the screenshots of the experiment showing the four goal configurations $\boldsymbol{q}_{goal,j}$ computed by the NSPG.

At the end of the $i$-th planned trajectory, the drilling procedure starts detecting all the $N_{DP}^{(i)}$ visible drilling points $\mathcal{F}_{DP,j}^{(i)}$ relative to the $i$-th drilling area. Then the robot starts drilling sequentially all the drilling points first aligning the drill with the surface normal, and then approaching the wall with the visual servoing based controller described in Sec. 5.4.3 and shown in Fig. 19.

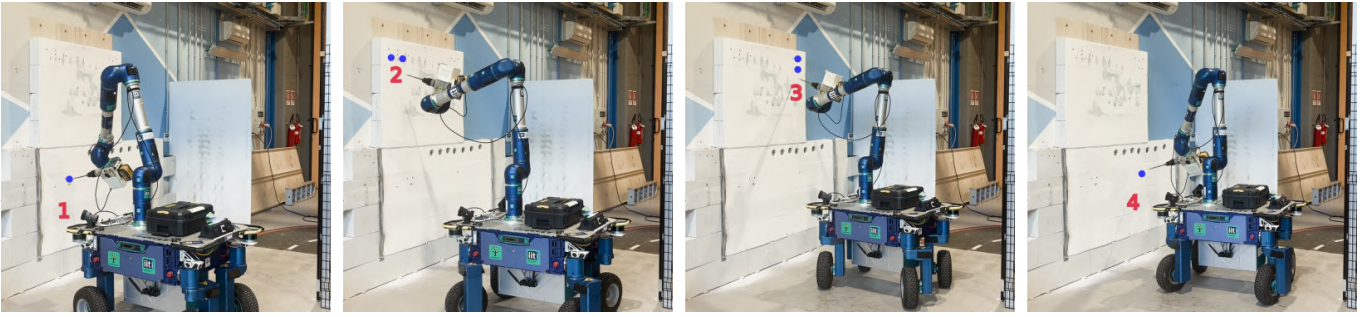

**FIGURE 18** Screenshots from the experiments showing the goal configurations found by the NSPG and used by the planner to drive the robot in front of the four drilling areas.

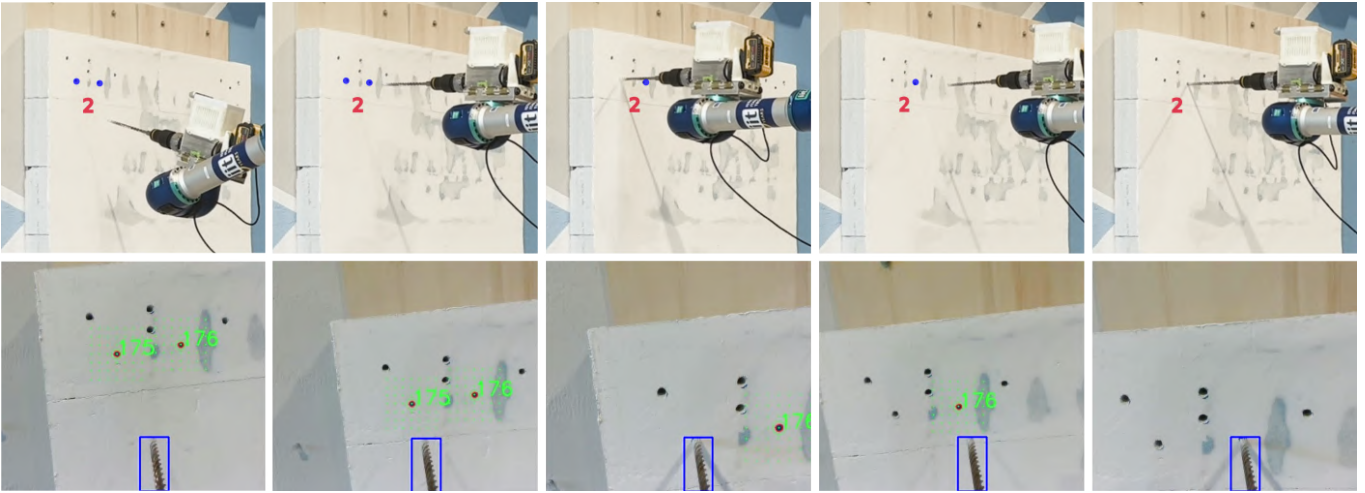

**FIGURE 19** Sequence of actions followed to drill the two drilling points $\mathcal{F}_{DP,1}^{(2)}$ and $\mathcal{F}_{DP,2}^{(2)}$ relative to the 2nd drilling area. The bottom row collects screenshots taken from the drill camera that detects the two drilling points labeled with numbers 175 and 176.

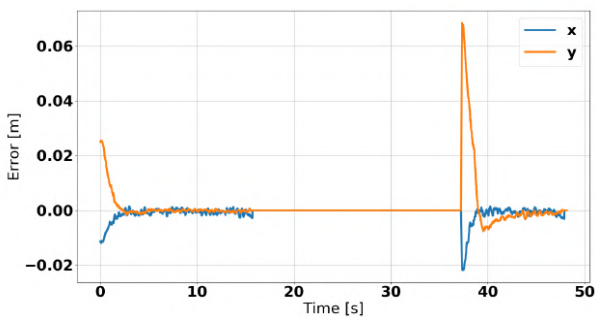

**FIGURE 20** Cartesian error on the x and y axis during the closed-loop drill of the two drilling points in Fig. 19.

The experiments regarding the second scenario has been carried out using morphology A. Screenshots are collected in Fig. 21. The user draws two blue dots on a wall close enough to be enclosed in a single drilling area. Then, the drilling area is manually selected from the BIM model trying to align in the most precise way possible the drilling area frame with the actual position of the dots just drawn. It is important to note that BIM models can encode specific construction site features, such as drilling areas, with high accuracy. Therefore, we expect the robot to navigate precisely to the designated drilling area and perform the required tasks autonomously, without the need for this manual alignment process, minimizing navigation errors from incorrect area selection. After selecting the drilling area, the behavior tree (Fig. 11) generates the corresponding drilling area frame $\mathcal{F}_{DA,0}$ and a Cartesian goal position $< x, y, \theta >$. This information is sent to the autonomous navigation system, which guides the robot to the specified drilling area. Once the robot reaches the wall, the whole-body planning module is activated after the NSPG computes the goal configuration $q_{goal}$, aligning the drill camera with the surface normal relative to the drilling area frame. The robot then initiates the closed-loop drilling process, executing a drill for each detected point.

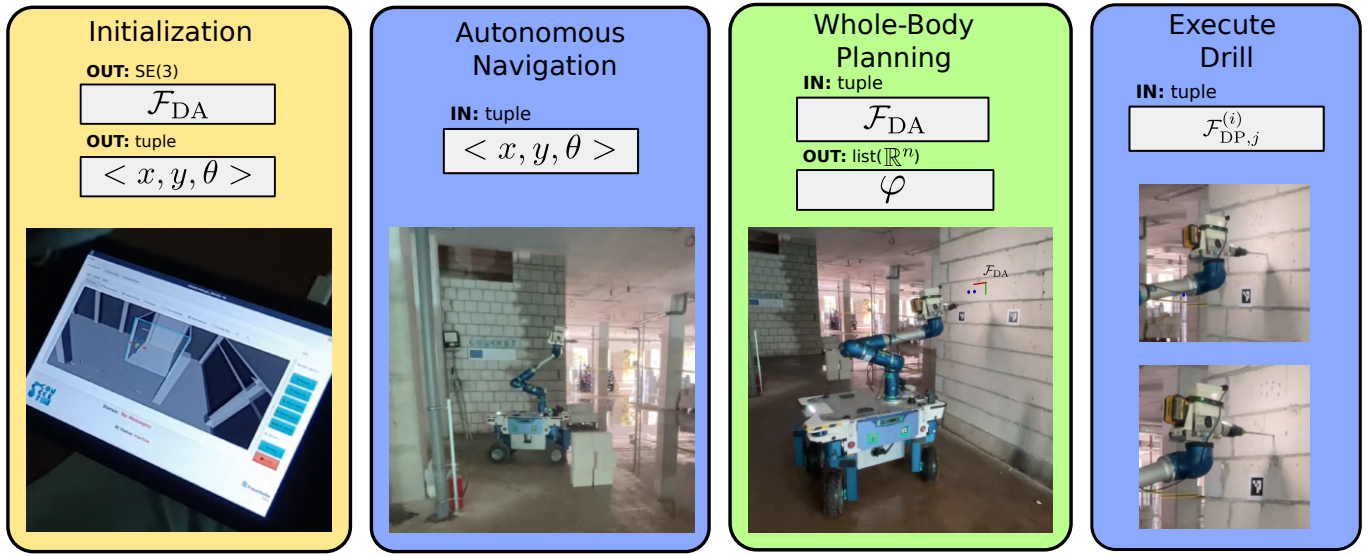

**FIGURE 21** Screenshots from the autonomous drilling task. From left to right: the behavior tree is initialized selecting the drilling area on the BIM and computing the drilling area frame $\mathcal{F}_{DA}$ and the goal Cartesian pose; the autonomous navigation stack starts to reach the goal pose; the whole-body planner aligns the drill camera frame with the drilling area frame; the robot drills the two blue makers detected.

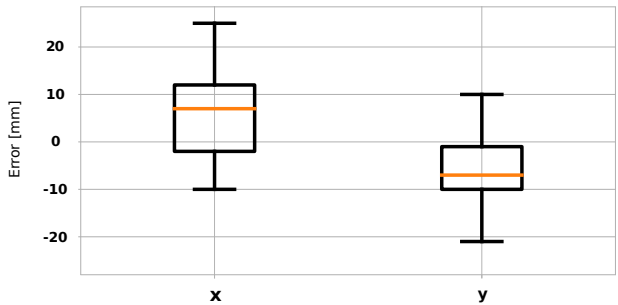

**FIGURE 22** Boxplot of the $x$ and $y$ error from the drill pattern experiments

Figure 20 illustrates the trend of the Cartesian error along the $x$ and $y$ axes during the closed-loop approach to the two drilling points depicted in Fig. 19. Initially, the closed-loop controller reduces the alignment error to near zero within approximately 2 s, maintaining a low error throughout the approach until the blue marker corresponding to $\mathcal{F}_{DP,1}^{(2)}$ remains detectable. The second phase of the closed-loop approach begins with a larger initial error, which the controller corrects, though with a slight overshoot occurring around 40 s.

For the third scenario, we tested the robot's performance drilling several vertical and horizontal patterns at different height. In this case, by detecting the ArUco pose from RGB data, the drilling point poses $\mathcal{F}_{DP,j}^{(i)}$ are automatically computed by setting the center, width, direction, and number of points of each pattern from the ArUco pose. Thus, the drills are not performed using the closed-loop visual servoing and the precision of the task only relies on the kinematic performance of the robot.

After the detection of the drilling area from the ArUco marker, and the computation of the drilling points frames, the planning pipeline follows the same steps as the previous scenarios. First, the whole-body motion planner that drives the robot with the drill-bit aligned with the ArUco marker. Then, the robot performs the drill tasks replacing the closed-loop visual servoing with a simple motion along the axis of the drillbit. The robot's performance is assessed by the error on the $x$ and $y$ axis between the desired drilling position and the actual one. We carried out a total of twenty-two experiments performing 60 cm wide horizontal patterns at different heights ranging from 0.35 m to 2.90 m. When drilling at 2.90 m, we use morphology C and the target height is achievable due to the base height, which is approximately 0.75 m.

## 6.2 | Spraying and Sanding

For this task, due to the larger surface area to be processed, we used the same morphology B described before installing the Spraying Tool End-Effector module instead of the Drill End-Effector. As stated before, the total length of the arm is about 2.15 m, mounting an end-effector of 1.8 kg. In this scenario, the robot had to apply an insulation layer on a wall, following the scheme described in Sec.5.7. After positioning the robot in front of the wall, the user highlights the area of interest by directly interacting with the GUI. Right after receiving the user's instructions, the robot moves at a specified distance and orientation from the wall and executes the task vertically spanning over the whole area.

We tested the performance of the robot executing the same task twice on two adjacent areas of the same wall. Specifically, the user selects an area of $w = 0.7\,\text{m} \times h = 0.8\,\text{m}$ with $w$ and $h$ being the width and the height of the selected rectangle. Since the width of each vertical line painted by the robot will depend on the distance of the end-effector from the wall, we heuristically set a distance of $0.3\,\text{m}$ in order to have a width for each painted line of $w_l = 0.1\,\text{m}$. In this way, the robot automatically computes the number of coats of insulating material necessary to cover the selected area by dividing the width of the rectangle times the width of each coat.

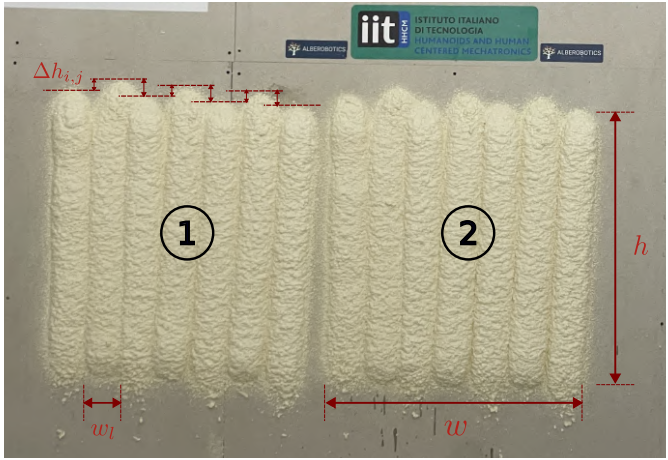

**FIGURE 23** Two adjacent areas of $0.8\,\text{m} \times 0.7\,\text{m}$ where insulating material has been applied by the CONCERT robot.

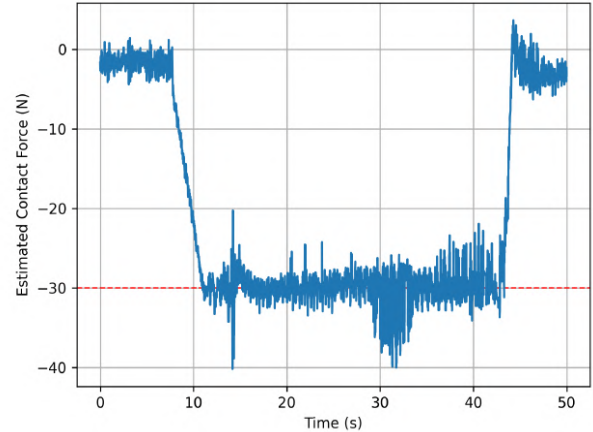

**FIGURE 24** Plot of the contact force between sanding end-effector and wall estimated from joint torques.

**TABLE 2** Numerical results from the plastering use-case

| | | | | | | |
|---|---|---|---|---|---|---|
| ① | $w_{l1}$ | 100 | $h_1$ | 800 | | |
| | $w_{l2}$ | 100 | $h_2$ | 810 | $\Delta h_{1,2}$ | 40 |
| | $w_{l3}$ | 97 | $h_3$ | 800 | $\Delta h_{2,3}$ | -50 |
| | $w_{l4}$ | 100 | $h_4$ | 800 | $\Delta h_{3,4}$ | 40 |
| | $w_{l5}$ | 100 | $h_5$ | 800 | $\Delta h_{4,5}$ | -40 |
| | $w_{l6}$ | 100 | $h_6$ | 800 | $\Delta h_{5,6}$ | 40 |
| | $w_{l7}$ | 96 | $h_7$ | 800 | $\Delta h_{6,7}$ | -40 |
| ② | $w_{l1}$ | 100 | $h_1$ | 800 | | |
| | $w_{l2}$ | 100 | $h_2$ | 800 | $\Delta h_{1,2}$ | 30 |
| | $w_{l3}$ | 97 | $h_3$ | 800 | $\Delta h_{2,3}$ | -30 |
| | $w_{l4}$ | 100 | $h_4$ | 800 | $\Delta h_{3,4}$ | 30 |
| | $w_{l5}$ | 100 | $h_5$ | 800 | $\Delta h_{4,5}$ | -25 |
| | $w_{l6}$ | 100 | $h_6$ | 800 | $\Delta h_{5,6}$ | 30 |
| | $w_{l7}$ | 100 | $h_7$ | 800 | $\Delta h_{6,7}$ | -35 |

The obtained result is shown in Fig.23 and in Table 2. Visually, the robot successfully completes the task by uniformly applying the insulating layer over the designated area of the wall. Additionally, the dimensions of each coat are highly consistent, indicating that the robot maintains the predefined distance from the wall throughout the entire process. However, as observed from the $\Delta h_{ij}$ measurements, there is a systematic error of a few centimeters between the starting points of adjacent coats. Since the insulating layer is applied by moving the end-effector in an up-and-down motion, this error may arise from the tool's velocity direction being equal and opposite to the gravity vector for adjacent coats. This phenomenon will be further examined in future work.

For the sanding task, we used the same arm as for the spraying task, removing the second Passive Link Module since the length of the Sanding Tool Module is enough to reach the whole workspace. Indeed, in this configuration, the total length of the robotic arm is around $2.10\,\text{m}$, comparable with the spraying case, carrying a tool with a mass of $2.90\,\text{kg}$. The validation of the sanding task was done empirically instead, as the quality of the final result is usually evaluated in a visual and tactile way. The contact force between the sander tool and the wall estimated from the joint torques is shown in Figure 24. For this sanding task, the user had selected a target contact force of $30\,\text{N}$. At the beginning, the robot approaches the wall linearly increasing the force applied on it. When the force setpoint is reached, the sanding task can start. We can see in the picture how the contact force is regulated (as explained in Section 5.7) during the motion to stay at $30\,\text{N}$. The spikes visible from the plot are due to holes and mounds left during plastering (performed by an inexperienced human operator) and in general due to imperfections on the wall. The estimated force is kept in a $10\,\text{N}$ boundary of the setpoint. In Figure 25 on the left we can see some defects on the plastered wall highlighted in green, while on the right the sanded area is highlighted in red.

# 7 | CONCLUSIONS

This paper presented CONCERT, a fully reconfigurable modular robotic platform for on-site operations in construction sites. Thanks to its versatility and the high power density of its motors, CONCERT can support human activities in collaborative tasks, but also operate autonomously to execute highly demanding tasks such as drilling, sanding, and plastering. To simplify the usage and reconfiguration of its kino-dynamic structure, we designed a software architecture, based on our middleware XBot2, which exploits topology recognition and automatic URDF generation to dynamically reconfigure the robot depending on the mounted modules. CONCERT has been validated performing a list of tasks typical of a construction site: collaborative transportation of high volume and heavy objects with a human operator, autonomous and semi-autonomous drilling in a height range from $0.35$ m to $2.90$ m, sanding, and plastering of insulating material. For each use-case, we used a robot configuration suited for the task execution, switching between a set of end-effectors controlled with the same software architecture we employ in the robot. This validation confirmed the high versatility, precision, and power that characterize CONCERT, being able to fulfill all the tasks both in manual and (semi-)autonomous way.

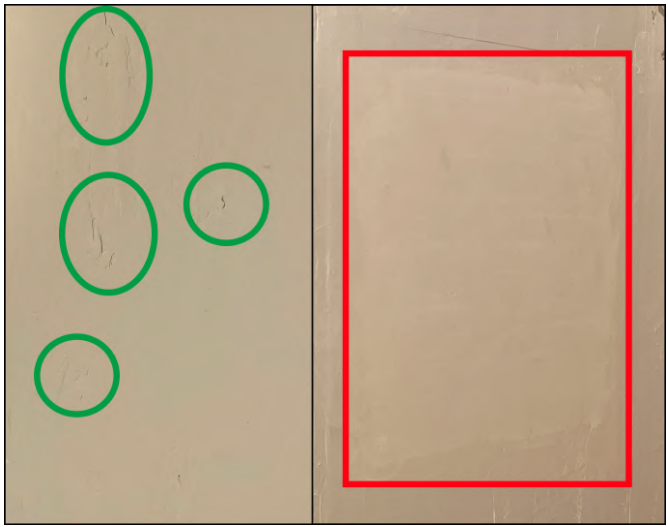

**FIGURE 25** A section of the wall used for evaluation of the sanding task. On the left imperfections on the wall before sanding are highlighted in green, on the right the sanded area is highlighted in red.

The experimental validation highlighted some systematic errors in different regions of the workspace that depend on the manual mounting of the unit modules and their mechanical interfaces. Indeed, despite the precise manufacture of the mechanical component, small backlashes lead to bigger errors the longer is the kinematic chain. To address this issue, we are designing a calibration setup to correct the Cartesian error in different areas of the workspace. The calibration setup is made by a metal plate with holes in known locations. By commanding the robot to move in these reference poses, and by compensating the Cartesian error between the actual Cartesian pose reached and the reference one, we can improve the performance in all the use-cases.

Future work will also focus on automatizing and optimizing the robot's morphology depending on the specific requirement of each task, taking advantage of some unit modules that have not been exploited so far, like the $45$ deg module or the dual arm configuration. Further, the usability of the robot by end-users should be addressed to test the easiness and reliability of the CONCERT robot to be employed in real scenarios by people who are not familiar with robotics and robot development. Preliminary results in this direction have already been recently presented in Lei et al. (2024), which proposes an algorithm for the optimization of the robot morphology minimizing the joint torques generated during the execution of a drilling task in pre-defined locations.

# 8 | ACKNOWLEDGEMENTS

□

# APPENDIX

# A  ETHERCAT NETWORK TOPOLOGY RECONSTRUCTION

The EtherCAT standard functioning, for what regards the network topology, will be briefly recalled here. In order to reconstruct the robot topology we assume that for each robot module, the embedded slave device includes at least one EtherCAT Slave Controller (ESC) chip.

Figure A1 illustrates the daisy chaining of two ESC chips complying with the EtherCAT standard. Only 4-port slaves (Port 0-3) will be considered. The EtherCAT network is an open ring and when the EtherCAT master inserts a data telegram into the ring, it will arrive to a slave through port 0. In an ESC, Port 0 has a special function as a so-called upstream port since it always points towards the network master. In each chip, an EtherCAT Processor Unit (EPU) is present which receives and adds data to the telegram. The telegram advances to the next port when a port is closed; open ports pass the telegram towards a downstream slave connected to them until it is received back through this port and forwarded to the next one. Finally, the telegram arrives back at port 0 and therefore to the master.

Any port of the ESC is automatically opened or closed when a communication link on that port becomes active or inactive. Each ESC exhibits a register that holds the open/closed state of each port. When the network is scanned, for each ESC, the EPU adds the value of this register on the data telegram. The order in which this data is inserted determines also the position of the slave in the EtherCAT ring, which is assigned by the master when it receives back the complete telegram. By accessing this register for each ESC and knowing the position of each ESC in the ring, it is therefore possible to reconstruct the parent/child relationship of the slaves and to infer the EtherCAT network topology.

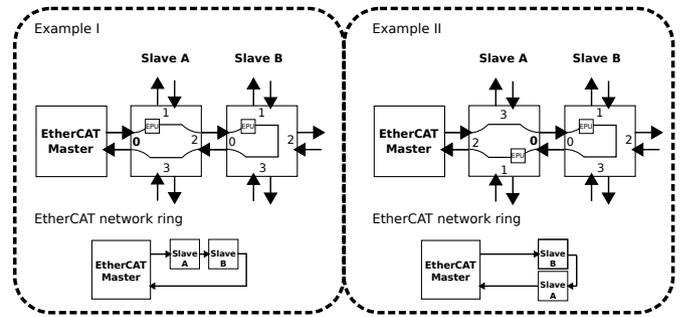

**FIGURE A1**  Illustration of how the EtherCAT data telegram travels from the master to the slaves through their ports. Ports are open (or closed) when a physical connection between two slaves is realized (or severed). The order in which EPUs are encountered, causes the master to see a different actual topology for Example I and II, where slave A was connected upside down, violating the EtherCAT standard. Not having port 0 as its upstream port, in Example II the EPU of slave A is encountered after slave B, which would result in an incorrect topology reconstruction.

The network implementation allows connecting slaves in *apparent topologies* that are bus, tree- or star-shaped, resulting in robots with series or tree-like kinematic chains. However, an EtherCAT *actual network topology* is always an open ring where only the organization of the ports in the ESC makes the apparent network topology – and thereby the robot's physical topology – appear differently. The master sees all slaves lined up in a certain order on this ring.

As an example of how the data is transmitted, let's consider we have a network with two EtherCAT slaves: A and B, as in Example I of Figure A1. Slave B is connected to port 2 of slave A. When scanning the network:

1. the master will pass the telegram to slave A via its upstream port 0
2. the telegram will be processed by the EPU of slave A
3. since no slave is connected to port 1, the port will result as closed letting the telegram proceed to the next port of slave A
4. port 2 will be open since the slave B is connected via this port, so the telegram will be passed to slave B
5. the telegram will be received by the slave B through its upstream port 0 and since ports 1, 2, and 3 are all closed, after being processed by the EPU will be transmitted back to port 0
6. the telegram comes back to slave A from its port 2 and since port 3 is closed goes back to port 0, where is sent back to the master.

In this case, the order of the slaves in the ring will be slave A first and slave B second. In Example II instead, slave A is connected "upside down" and port 0 is not the upstream port, violating the EtherCAT protocol. This leads to the telegram arriving at the EPU of slave A after that of slave B and results in having a different actual topology from the one in Example I. In this case, the order will be: slave B first, slave A second. Although Example II would not be allowed by the EtherCAT protocol, it represents a situation that can arise in reality, if the ESCs are connected in the wrong way. Since the EtherCAT master has no way of detecting that this might have happened, it is the task of the user to ensure the correct connection between the ESCs and in our case between the modules. If port 0 would not be set as upstream port the EtherCAT topology reconstruction would be wrong and that would lead to an erroneous representation of the robot's physical topology.

In our case, each port of the ESC is associated with one EMI of the module, and each established mechanical connection corresponds to an open communication link. This means each module will need to be connected with the EMI associated with port 0 of his ESC looking upstream. This entails that a module can be connected to its predecessor (parent) module in only one way, preventing the possibility of attaching a module upside down. Because of this the EMI associated with the "upstream port" is static and we'll refer to this as "upstream EMI".

## A.1 Topology reconstruction algorithm

The EtherCAT master can then reconstruct the apparent network topology from the network ring by looping over all slaves on the ring. Each slave has a parent slave in the tree-like apparent network topology. This parent is a precursor on the network ring, but not necessarily the direct neighbor of the slave. Considering the EtherCAT network rules, it is possible, by looking at the open ports of each slave on the ring, to determine its parent, as implemented in the Simple Open EtherCAT Master [§§]. The algorithm to find the parent position given the position of the slave in the ring (*slave_pos*) is shown in Algorithm 1.

The function GetOpenPorts(*slave_idx*) would be retrieving the value of the register of open ports for a specific slave at position *slave_idx*. This function will return a list of the ports that are open for that slave, which can take values from $\{0, 1, 2, 3\}$. If a slave has two children attached to port 2 and 3 the function will return $[0, 2, 3]$ since zero is always open being the downstream port.

To completely reconstruct the graph representing the network topology for each slave we need to know not only the parent but also to which port of the parent the slave is connected. In this way, we can distinguish between robots with the same actual topology but different apparent topology. Algorithm 2 describes the procedure that can be used during the scan of the network to reconstruct the network topology.

Each slave information will be inserted in a Slave data structure containing: a code identifying the module associated with that slave (*module_identifier*), the position of the slave in the ring (*position*), the position of the parent (*parent_position*), the port of the parent to which

---
**Algorithm 1** Algorithm to determine position of parent slave in the EtherCAT ring

---
1: **function** FindParentPosition(*slave_pos*)
2:     *parent_pos* ← 0
3:     **if** *slave_pos* > 1 **then**
4:         *candidate_parent* ← *slave_pos* − 1
5:         *topology_counter* ← 0
6:         **while** *candidate_parent* > 0 **do**
7:             *open_ports* ← GetOpenPorts(*candidate_parent*)
                ▷ Retrieve open ports for the slave at the required position
8:             *num_open_ports* ← *length*(*open_ports*)
9:             **if** *num_open_ports* == 1 **then**          ▷ end point
10:                 *topology_counter* ← *topology_counter* − 1
11:             **else if** *num_open_ports* == 2 **then**     ▷ link point
12:                 *topology_counter* ← *topology_counter*
13:             **else if** *num_open_ports* == 3 **then**     ▷ split point
14:                 *topology_counter* ← *topology_counter* + 1
15:             **else if** *num_open_ports* == 4 **then**     ▷ cross point
16:                 *topology_counter* ← *topology_counter* + 2
17:             **end if**
18:             **if** (*topology_counter* ≥ 0 ∧ *num_open_ports* > 1)
                  ∨ (*candidate_parent* == 1) **then**
19:                 *parent_pos* ← *candidate_parent*      ▷ Parent found
20:                 *candidate_parent* ← 0
21:             **end if**
22:             *candidate_parent* ← *candidate_parent* − 1
23:         **end while**
24:     **end if**
25:     **return** *parent_pos*
26: **end function**

---

the slave is connected (*parent_port*) and a stack that at the beginning of the scan contains the open ports of the slave (*free_ports_stack*). This stack is temporary: the algorithm will use it to assign the correct port of the parent to which a slave is connected. It will be empty at the end of the scan. The function GetModuleIdentifier serves to retrieve the *module_identifier* stored in the microprocessor of the slave device. This parameter is not useful for the network topology reconstruction, but it will be used to reconstruct the physical robot topology. All the Slave instances will be inserted into a list, where the indexes will represent the position of a certain slave in the ring.

This list just created by scanning the network will contain the same information as a graph representing the parent/child relationship of the slaves. The result of this step is a graph structure $\chi$ representing the apparent network topology with the slaves being the nodes and the connection among the ports being the edges of the graph.

---

---

**Algorithm 2** Algorithm to reconstruct the network topology

---

1: $n$ is the number of slaves in the network

2: $\mathcal{P} := \{1, 2, 3\}$ is the set of ports each slave can have open if a child slave is connected to it. Port 0 is excluded as it will always be open

3: **struct** Slave

4:      uuid *module_identifier*

5:      uint *position* $\in [1, n]$

6:      uint *parent_position* $\in [1, n]$

7:      uint *parent_port* $\in \mathcal{P}$

8:      stack *free_ports_stack*, where $\forall p \in \textit{free\_ports\_stack} : p \in \mathcal{P}$

9: **end struct**

10: **function** NetworkTopologyRecognition

11:      Initialize *slave_list* as a list of Slave elements of length $n$.

12:      **for** $i \leftarrow 1, n$ **do**

13:          *parent_pos* $\leftarrow$ FindParentPosition($i$)      ▷ see Algorithm 1

14:          **if** *parent_pos*! $= 0$ **then**

15:              *parent_slave* $\leftarrow$ *slave_list*[*parent_pos*]

16:          **else**

17:              *parent_slave* $\leftarrow$ *null*

18:          **end if**

19:          *slave_list*[$i$].*module_identifier* $\leftarrow$ GetModuleIdentifier($i$)

20:          *slave_list*[$i$].*position* $\leftarrow i$

21:          *slave_list*[$i$].*parent_position* $\leftarrow$ *parent_pos*

22:          *open_ports* $\leftarrow$ GetOpenPorts($i$)

23:          $j \leftarrow$ *length*(*open_ports*)

24:          **while** $j > 0$ **do**

25:              push *open_ports*[$j$] to *slave_list*[$i$].*free_ports_stack*

26:              $j \leftarrow j - 1$

27:          **end while**

28:          **if** *parent_pos*! $= 0$ **then**

29:              pop *top* from *parent_slave.free_ports_stack*

30:              *slave_list*[$i$].*parent_port* $\leftarrow$ *top*

31:          **else**

32:              *slave_list*[$i$].*parent_port* $\leftarrow$ *null*

33:          **end if**

34:      **end for**

35:      **return** *slave_list*

36: **end function**

---